\documentclass{article}
\usepackage{ijcai22}

\usepackage{times}
\usepackage{soul}
\usepackage{url}
\usepackage[hidelinks]{hyperref}
\usepackage[utf8]{inputenc}
\usepackage[small]{caption}
\usepackage{graphicx}
\usepackage{amsmath}
\usepackage{amsthm}
\usepackage{booktabs}
\usepackage{algorithm}
\usepackage{algorithmic}
\usepackage{amsfonts,bm}
\usepackage{subfigure}
\usepackage{enumitem}
\usepackage{multirow,color}
\newcommand{\model}{{{\tt L2E}}}
\allowdisplaybreaks

\newtheorem{theorem}{Theorem}
\newtheorem{lemma}{Lemma}
\newtheorem{corollary}{Corollary}
\newtheorem{definition}{Definition}

\title{A Unified Meta-Learning Framework for Dynamic Transfer Learning}

\author{
    Jun Wu
    \and
    Jingrui He
    \affiliations
    University of Illinois Urbana-Champaign
    \emails
    \{junwu3, jingrui\}@illinois.edu
}

\begin{document}

\maketitle

\begin{abstract}
  Transfer learning refers to the transfer of knowledge or information from a relevant source task to a target task. However, most existing works assume both tasks are sampled from a stationary task distribution, thereby leading to the sub-optimal performance for dynamic tasks drawn from a non-stationary task distribution in real scenarios. To bridge this gap, in this paper, we study a more realistic and challenging transfer learning setting with dynamic tasks, i.e., source and target tasks are continuously evolving over time. We theoretically show that the expected error on the dynamic target task can be tightly bounded in terms of source knowledge and consecutive distribution discrepancy across tasks. This result motivates us to propose a generic meta-learning framework \model{} for modeling the knowledge transferability on dynamic tasks. It is centered around a task-guided meta-learning problem with a group of meta-pairs of tasks, based on which we are able to learn the prior model initialization for fast adaptation on the newest target task. \model{} enjoys the following properties: (1) effective knowledge transferability across dynamic tasks; (2) fast adaptation to the new target task; (3) mitigation of catastrophic forgetting on historical target tasks; and (4) flexibility in incorporating any existing static transfer learning algorithms. Extensive experiments on various image data sets demonstrate the effectiveness of the proposed \model{} framework.
\end{abstract}

\section{Introduction}
Transfer learning~\cite{pan2009survey} aims to leverage the knowledge of a source task to improve the generalization performance of a learning algorithm on a target task. The knowledge transferability across tasks can be theoretically guaranteed under mild assumptions, even when no labeled training examples are available in any target task~\cite{ben2010theory,ghifary2016scatter,pmlr-v139-acuna21a}. One key assumption is that source and target tasks are sampled from a stationary task distribution. The resulting relatedness between tasks allows transferring knowledge from a source task with adequate labeled data to a target task with little or no labeled data. However, the learning task might be evolving over time~\cite{mohri2012new} in real scenarios. For example, the data distribution of clothes images in Amazon is changing over the years due to the varying fashion trend~\cite{al2017fashion}, thus resulting in a time-evolving image recognition task. Another example is the IMDb's film rating system where the rating scores of a film change in different time periods due to the dynamic user preference~\cite{rafailidis2015modeling}, thus leading to a time-evolving film recommendation task. Such application scenarios would challenge the conventional static transfer learning algorithms~\cite{pan2009survey} due to the dynamic task relatedness.

\begin{figure}
    \centering
    \includegraphics[width=0.48\textwidth]{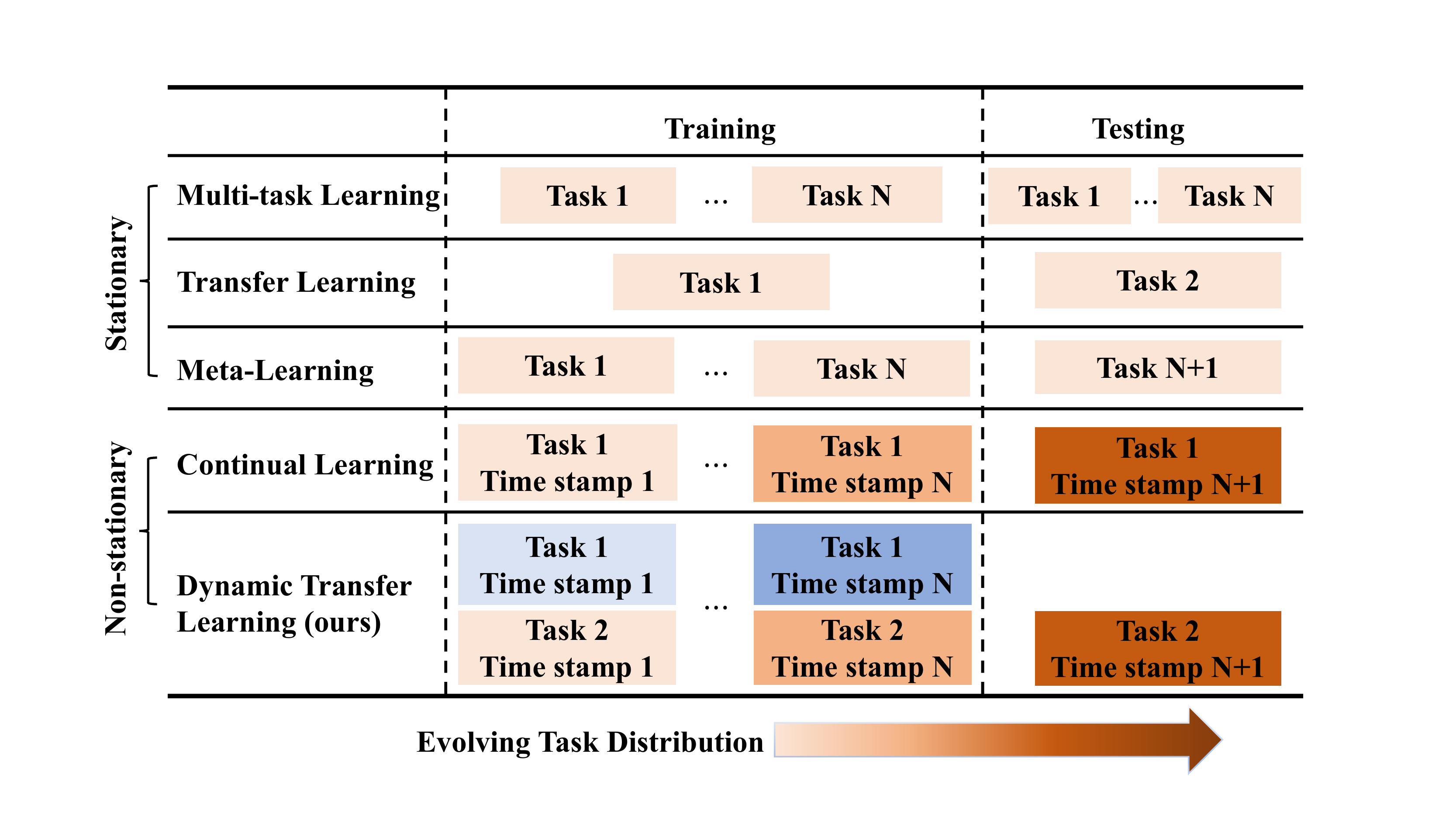}
    \caption{Illustration of transfer learning on dynamic tasks}
    \label{fig:problem_setting}
\end{figure}

Recent works~\cite{hoffman2014continuous,liu2020learning,wang2020continuously,kumar2020understanding} have studied continuous transfer learning with a static source task and a time evolving target task. They revealed that the prediction function on the newest target task can be learned by aggregating the knowledge from the labeled source data and historical unlabeled target data. Nevertheless, in real scenarios, both source and target tasks could be changing over time. In this case, those works will lead to the sub-optimal solution due to the under-explored source knowledge. To the best of our knowledge, very little effort has been devoted to modeling the knowledge transferability from a labeled dynamic source task to an unlabeled dynamic target task.

To bridge this gap, in this paper, we study the dynamic transfer learning problem with dynamic source and target tasks sampled from a non-stationary task distribution. As shown in Figure~\ref{fig:problem_setting}, conventional knowledge transfer problems focus on either static tasks (e.g., multi-task learning~\cite{Multic2_Yao_2017,zhou2019multi}, transfer learning~\cite{pan2009survey} and meta-learning~\cite{finn2017model}) or one dynamic task (e.g., continual learning~\cite{li2017learning}), whereas we aim to transfer the knowledge from a dynamic source task to a dynamic target task. More specifically, we focus on the learning scenario where both source and target tasks always share the same class-label space (a.k.a. domain adaptation~\cite{zhang2019bridging}) at any time stamp. We are able to show that the generalization error bounds of dynamic transfer learning can be derived under the following assumptions. First, the class labels of the source task are available at any time stamp. Second, the source and target tasks are related at the initial time stamp. Third, the data distributions of both source and target tasks are continuously changing over time. The theoretical results indicate that the target error is bounded in terms of flexible domain discrepancy measures (e.g., $\mathcal{H}$-divergence~\cite{ben2010theory}, discrepancy distance~\cite{mansour2009domain}, $f$-divergence~\cite{pmlr-v139-acuna21a}, etc.) across tasks and across time stamps. This motivates us to propose a generic meta-learning framework \model\ for dynamic transfer learning. It reformulates the dynamic source and target tasks into a set of meta-pairs of consecutive tasks, and then learns the prior model initialization for fast adaptation on the newest target task. The effectiveness of \model\ is empirically verified on various image data sets. 
The main contributions of this paper are summarized as follows:
\begin{itemize}
    \item We derive the error bounds for dynamic transfer learning with time-evolving source and target tasks.
    \item We propose a generic meta-learning framework (\model) for transfer learning on dynamic tasks by minimizing the error upper bounds with flexible divergence measures.
    \item Extensive experiments on public data sets confirm the effectiveness of our proposed \model{} framework.
\end{itemize}

The rest of the paper is organized as follows. We review the related work in Section~\ref{sec:related_work}, followed by our problem setting in Section~\ref{sec:preliminaries}. In Section~\ref{sec:algorithm}, we derive the error bounds of continuous transfer learning and then present the \model{} framework. The extensive experiments and discussion are provided in Section~\ref{sec:experiments}. Finally, we conclude the paper in Section~\ref{sec:conclusion}.

\section{Related Work}\label{sec:related_work}
\subsection{Transfer Learning}
Transfer learning~\cite{pan2009survey} improves the generalization performance of a learning algorithm under distribution shift. Most existing algorithms~\cite{zhang2019bridging,pmlr-v139-acuna21a,wu2021indirect} assume the relatedness of static source and target tasks in order to guarantee the success of knowledge transfer. The most related problem to our work is the meta-transfer learning~\cite{sun2019meta}, which learns to adapt to a set of few shot learning tasks sampled from a stationary task distribution. However, our work focuses on the non-stationary task distribution where a sequence of meta-pairs of consecutive tasks could be formulated for dynamic transfer learning.

\subsection{Continual Learning}
Continual learning aims to learn a model on a new task using knowledge from experienced tasks. Conventional continual learning algorithms~\cite{li2017learning} focused on mitigating the catastrophic forgetting when learning the prediction function on one time-evolving task. In contrast, continuous transfer learning~\cite{hoffman2014continuous,bobu2018adapting,wu2020continuous,liu2020learning,wang2020continuously,kumar2020understanding} transferred the knowledge from a labeled static source task to an unlabeled time-evolving target task. Our work would further extend it to the transfer learning setting with dynamic source and target tasks.

\subsection{Meta-Learning}
Meta-learning~\cite{finn2017model,fallah2020convergence}, or learning to learn, leverages the knowledge from a set of prior tasks for fast adaptation to unseen tasks. It assumes that all the tasks follow a stationary task distribution. More recently, it has been extended into the online learning setting~\cite{finn2019online} where a sequence of tasks are sampled from non-stationary task distributions. However, those works focused on improving the prediction performance with the accumulated data, whereas we aim to explore the knowledge transferability between dynamic source and target tasks.

\section{Problem Setting}\label{sec:preliminaries}
Let $\mathcal{X}$ and $\mathcal{Y}$ be the input feature space and output label space respectively. We consider the dynamic transfer learning problem\footnote{In this paper, we assume that all the tasks share the same output label space $\mathcal{Y}$ for simplicity. Besides, we use $\mathcal{D}_j^s$ to represent both the time-specific task (i.e., source task at the $j^{\text{th}}$ time stamp) and its data distribution (i.e., probability distribution of the source task at the $j^{\text{th}}$ time stamp over $\mathcal{X}\times \mathcal{Y}$) for notation simplicity.} with dynamic source task $\{\mathcal{D}_j^s\}_{j=1}^N$ and target task $\{\mathcal{D}_j^t\}_{j=1}^N$ with time stamp $j$. In this case, we assume that there are $m_j^s$ labeled training examples $D_{j}^s = \{(\mathbf{x}_{ij}^s, y_{ij}^s)\}_{i=1}^{m_{j}^s}$ in the $j^\text{th}$ source task and no labeled training examples in the target task. Let $m_j^t$ be the number of unlabeled training examples $D_j^t=\{\mathbf{x}_{ij}^t\}_{i=1}^{m_j^t}$ in the $j^\text{th}$ target task. Furthermore, each task $\mathcal{D}_j^s$ ($\mathcal{D}_j^t$) is associated with a task-specific labeling function $f_j^s$ ($f_j^t$). Let $\mathcal{H}$ be the hypothesis class on $\mathcal{X}$ where a hypothesis is a function $h:\mathcal{X}\rightarrow\mathcal{Y}$. $\mathcal{L}(\cdot, \cdot)$ is the loss function such that $\mathcal{L}: \mathcal{Y}\times\mathcal{Y}\rightarrow\mathbb{R}$. The expected classification error on the task $\mathcal{D}_j$ (either source or target) is defined as $\epsilon_j(h)=\mathbb{E}_{(\mathbf{x},y)\sim\mathcal{D}_j} [\mathcal{L}(h(\mathbf{x}), y)]$ for any $h\in \mathcal{H}$, and its empirical estimate is given by $\hat{\epsilon}_j(h) = \frac{1}{m_j}\sum_{i=1}^{m_j} \mathcal{L}(h(\mathbf{x}_{ij}), y_{ij}) $.

Formally, our dynamic transfer learning problem can be defined as follows.
\begin{definition} ({\bf Dynamic Transfer Learning})
Given labeled dynamic source tasks $\{\mathcal{D}_j^s\}_{j=1}^N$ and unlabeled dynamic target tasks $\{\mathcal{D}_j^t\}_{j=1}^N$, dynamic transfer learning aims to learn the prediction function for the newest target task $\mathcal{D}_{N+1}^t$ by leveraging the knowledge from historical source and target tasks.
\end{definition}

In dynamic transfer learning, we have the following mild assumptions. (1) The class labels of the source task are available at any time stamp. Specially, when the source task is static, it is naturally degenerated into a continuous transfer learning problem~\cite{hoffman2014continuous,bobu2018adapting,wu2020continuous,liu2020learning}. (2) The source and target tasks are related at the initial time stamp $j=1$. That is, those tasks might not be related in the following time stamps $j> 1$, as their data distributions can be evolving towards different patterns. (3) The data distributions of both source and target tasks are continuously changing over time. 

\section{The Proposed Framework}\label{sec:algorithm}
In this section, we first derive the error bounds for dynamic transfer learning, and then propose a generic meta-learning framework (\model).

\subsection{Error Bounds}
Following~\cite{ben2010theory}, we consider a binary classification problem with $\mathcal{Y}\in \{0, 1\}$ for simplicity. Before deriving the generalization error bound for a dynamic target task, we first introduce some basic concepts below. 

\begin{definition} ({\bf $L^1$-divergence}~\cite{ben2010theory})
The $L^1$-divergence between two distributions $\mathcal{D}$ and $\mathcal{D}'$ over $\mathcal{X}$ is defined as follows.
\begin{equation}
    d_1(\mathcal{D}, \mathcal{D}') := 2 \sup_{Q\in \mathcal{Q}} \left|\mathrm{Pr}_{\mathcal{D}}[Q] - \mathrm{Pr}_{\mathcal{D}'}[Q] \right|
\end{equation}
where $\mathcal{Q}$ is the set of measurable subsets under $\mathcal{D}$ and $\mathcal{D}'$. 
\end{definition}

\begin{definition} ({\bf $\mu$-admissibility})
A loss function $\mathcal{L}(\cdot,\cdot)$ is $\mu$-admissible if there exists $\mu>0$ such that for all $\mathbf{x}\in \mathcal{X}$, $y,y'\in \mathcal{Y}$ and $h,h'\in \mathcal{H}$, the following inequalities hold.
\begin{align*}
    \left| \mathcal{L}(h'(\mathbf{x}), y) - \mathcal{L}(h(\mathbf{x}), y) \right| &\leq \mu \left| h'(\mathbf{x}) - h(\mathbf{x}) \right| \\
    \left| \mathcal{L}(h(\mathbf{x}), y') - \mathcal{L}(h(\mathbf{x}), y) \right| &\leq \mu \left| y' - y \right|
\end{align*}
\end{definition}
For dynamic transfer learning, the following theorem states that the expected target error of the newest target task can be bounded in terms of historical source and target knowledge.

\begin{theorem}\label{continuous_transfer_learning_bound}
Assume that the loss function $\mathcal{L}(\cdot, \cdot)$ is $\mu$-admissible and obeys the triangle inequality. Given a class of functions $\mathcal{H}_{\mathcal{L}} = \{(\mathbf{x}, y) \mapsto \mathcal{L}(h(\mathbf{x}), y): h\in \mathcal{H} \}$, for any $\delta>0$ and $h\in \mathcal{H}$, with probability at least $1-\delta$, the expected error $\epsilon_{N+1}^t$ for the newest target task $\mathcal{D}_{N+1}^t$ is bounded by
\begin{align*}
    \epsilon_{{N+1}}^t(h) &\leq \frac{1}{2N} \sum_{j=1}^N \left(\hat{\epsilon}_j^s(h) + \hat{\epsilon}_j^t(h) \right) + \frac{N+2}{2} \left( \Tilde{d} + \Tilde{\lambda} \right) \\
    &\quad + \Tilde{\Re}(\mathcal{H}_{\mathcal{L}}) + \frac{\mu}{N}\sqrt{\frac{\log{\frac{1}{\delta}}}{2\Tilde{m}}}
\end{align*}
where $\Tilde{d} =\mu\cdot \max\big\{\max_{1\leq j\leq N-1}d_1(\mathcal{D}^s_j, \mathcal{D}^s_{j+1}), d_1(\mathcal{D}^s_1, \mathcal{D}^t_1),$ $\max_{1\leq j\leq N}d_1(\mathcal{D}^t_j, \mathcal{D}^t_{j+1}) \big\}$, $\Tilde{\lambda} =\mu\cdot \max\big\{\max_{1\leq j\leq N-1}$ $ \lambda_*(\mathcal{D}^s_j, \mathcal{D}^s_{j+1}), \lambda_*(\mathcal{D}^s_1, \mathcal{D}^t_1), \max_{1\leq j\leq N}\lambda_*(\mathcal{D}^t_j, \mathcal{D}^t_{j+1}) \big\}$ and $\lambda_*$ measures the difference of the labeling functions across task and across time stamps, i.e., $\lambda_*(\mathcal{D}^s_1, \mathcal{D}^t_1) = \min\{\mathbb{E}_{\mathcal{D}^s_1}[|f_1^s(\mathbf{x}) -f_1^t(\mathbf{x})|], \mathbb{E}_{\mathcal{D}^t_1}[|f_1^s(\mathbf{x}) -f_1^t(\mathbf{x})|]\}$. $\Tilde{\Re}(\mathcal{H}_{\mathcal{L}})$ is a Rademacher complexity term and $\Tilde{m}=$ $\sum_{j=1}^N(m_j^s+m_j^t)$ is the total number of training examples from source and historical target tasks.
\end{theorem}

This theorem reveals that the expected error on the newest target task is bounded in terms of (i) the empirical errors of historical source and target tasks; (ii) the maximum of the distribution discrepancy across tasks and across time stamps; (iii) the maximum of the labeling difference across tasks and across time stamps; and (iv) the average Rademacher complexity~\cite{mansour2009domain} of the class of functions $\mathcal{H}_{\mathcal{L}} = \{(\mathbf{x}, y) \mapsto \mathcal{L}(h(\mathbf{x}), y): h\in \mathcal{H} \}$ over all the tasks.

However, it has been observed that (1) $L^1$-divergence cannot be accurately estimated from finite samples of arbitrary distributions~\cite{ben2010theory}; and (2) the generalization error bound with $L^1$-divergence is not very tight because $L^1$-divergence involves all the measurable subsets over $\mathcal{X}$. Therefore, we would like to derive much tighter error bounds with existing domain divergence measures over either marginal feature space (e.g., $\mathcal{H}$-divergence~\cite{ben2010theory}, $f$-divergence~\cite{pmlr-v139-acuna21a} and Maximum Mean Discrepancy (MMD)~\cite{gretton2012kernel}) or joint feature and label space (e.g., $\mathcal{C}$-divergence~\cite{wu2020continuous}). It is notable~\cite{pmlr-v139-acuna21a} that the generic $f$-divergence subsumes many popular divergences, including Margin Disparity Discrepancy~\cite{zhang2019bridging}, Jensen-Shannon (JS) divergence, etc.

\begin{corollary}\label{col:marginal}
Assume that the loss function $\mathcal{L}(\cdot, \cdot)$ is $\mu$-admissible and symmetric (i.e., $\mathcal{L}(y_1, y_2) = \mathcal{L}(y_2, y_1) $ for $y_1,y_2\in \mathcal{Y}$), and obeys the triangle inequality. Then
\begin{enumerate}[label=(\alph*), leftmargin=*,noitemsep,nolistsep]
\item when using {\bf $f$-divergence}~\cite{pmlr-v139-acuna21a}\footnote{Note that we have similar error bounds when using other marginal domain discrepancy measures (e.g., $\mathcal{H}$-divergence~\cite{ben2010theory}, discrepancy distance~\cite{mansour2009domain}, or MMD~\cite{gretton2012kernel}), so we omit the details here (see Appendix for more illustration)}, denoted by $d_f$, the error bound of Theorem~\ref{continuous_transfer_learning_bound} holds with
\begin{align*}
    \Tilde{d} &= \max\big\{\max_{1\leq j\leq N-1}d_f(\mathcal{D}^s_j, \mathcal{D}^s_{j+1}), d_f(\mathcal{D}^s_1, \mathcal{D}^t_1), \\
    &\quad\quad\quad\quad \max_{1\leq j\leq N}d_f(\mathcal{D}^t_j, \mathcal{D}^t_{j+1}) \big\} \\
    \Tilde{\lambda} &= \max\big\{\max_{1\leq j\leq N-1} \lambda_*(\mathcal{D}^s_j, \mathcal{D}^s_{j+1}), \lambda_*(\mathcal{D}^s_1, \mathcal{D}^t_1), \\
    &\quad\quad\quad\quad \max_{1\leq j\leq N}\lambda_*(\mathcal{D}^t_j, \mathcal{D}^t_{j+1}) \big\}
\end{align*}
where $\lambda_*(\mathcal{D}^s_1, \mathcal{D}^t_1) = \min_{h\in \mathcal{H}} \epsilon_1^s(h) + \epsilon_1^t(h)$.

\item when using {\bf $\mathcal{C}$-divergence}~\cite{wu2020continuous} (measuring the distribution discrepancy over joint data distribution on $\mathcal{X}\times \mathcal{Y}$), denoted by $d_{\mathcal{C}}$, the error bound of Theorem~\ref{continuous_transfer_learning_bound} holds with
\begin{align*}
    \Tilde{d} &=\mu\cdot \max\big\{\max_{1\leq j\leq N-1}d_{\mathcal{C}}(\mathcal{D}^s_j, \mathcal{D}^s_{j+1}), d_{\mathcal{C}}(\mathcal{D}^s_1, \mathcal{D}^t_1),\\ 
    &\quad\quad\quad\quad\quad \max_{1\leq j\leq N}d_{\mathcal{C}}(\mathcal{D}^t_j, \mathcal{D}^t_{j+1}) \big\} \\
    \Tilde{\lambda} &= 0
\end{align*}
\end{enumerate}
\end{corollary}

These theoretical results motivate us to develop a dynamic transfer learning framework by empirically minimizing the error bounds with flexible domain discrepancy measures.

\subsection{\model\ Framework}
\begin{figure}
    \centering
    \includegraphics[width=0.47\textwidth]{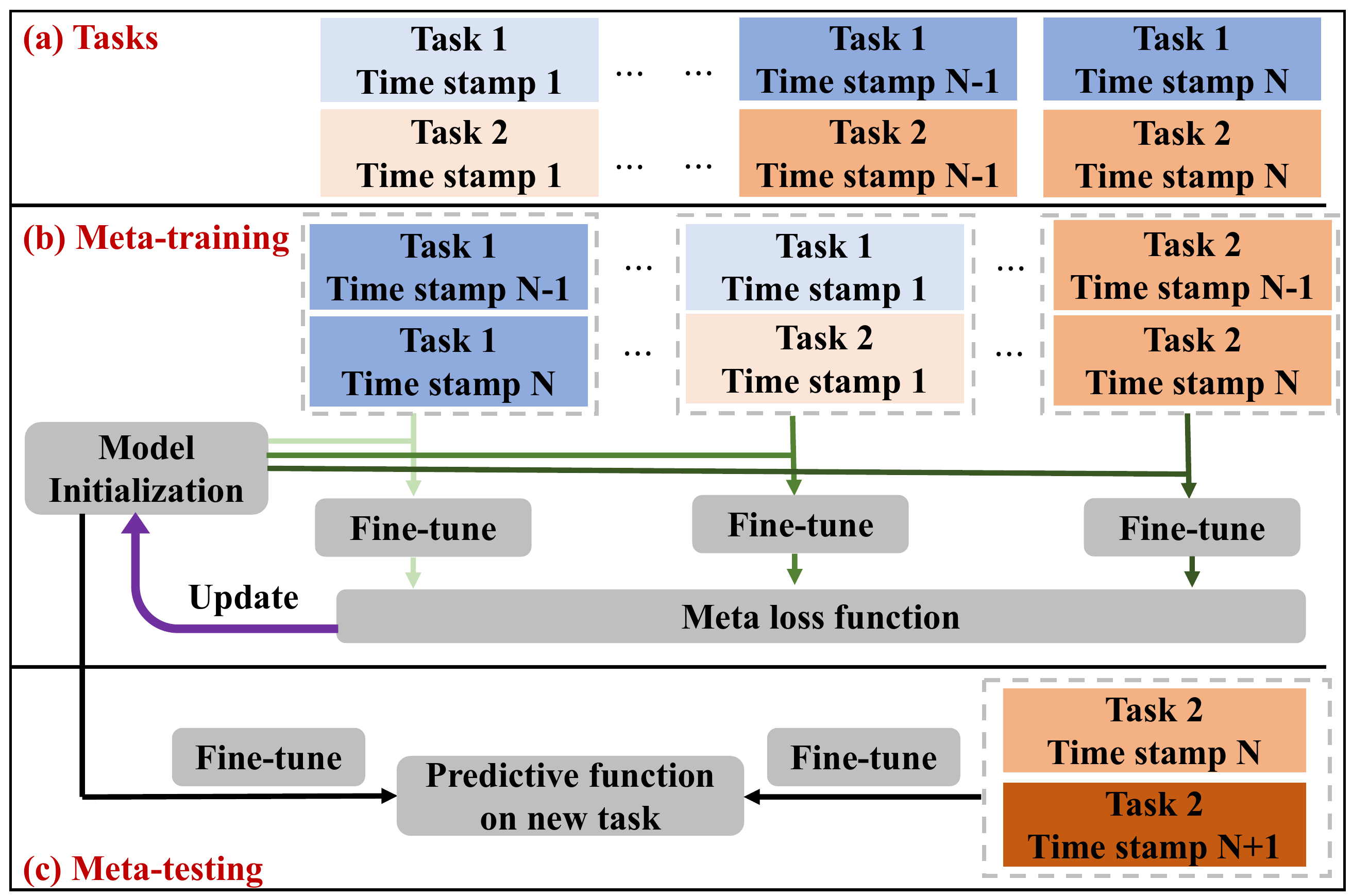}
    \caption{Illustration of our proposed \model{} framework}
    \label{fig:framework}
\end{figure}

Following~\cite{ben2010theory}, a typical transfer learning paradigm on static source and target tasks aims to minimize the static error bound involving empirical source classification error and domain discrepancy across tasks as follows.
\begin{equation}\label{eq:base_model}
    \min_{\theta} J(\theta) = \hat{\epsilon}_{s}(\theta) + \gamma \cdot \hat{d}(\mathcal{D}^s, \mathcal{D}^t; \theta)
\end{equation}
where $\theta$ is the trainable parameters and $\gamma$ is a trade-off parameter between the empirical source error and the domain discrepancy. The second term $\hat{d}(\mathcal{D}^s, \mathcal{D}^t; \theta)$ aims to match the data distribution of source and target tasks by learning the domain-invariant latent representation for every input example. Then the predictive function learned by the first term $\hat{\epsilon}_{s}(\theta)$ on the source task could be applied to the target task directly. In this case, existing works~\cite{ganin2016domain,pmlr-v139-acuna21a} have attempted to instantiate different domain discrepancy measures. Nevertheless, when the source and target tasks are evolving over time, it might be sub-optimal when directly transferring the labeled source task to the newest unlabeled target task. That is because the success of knowledge transfer cannot be guaranteed if the task relatedness becomes weaker over time~\cite{rosenstein2005transfer}.

In this paper, we propose a generic meta-learning framework \model{} for transferring the knowledge from a dynamic source task to a dynamic target task. It leverages the knowledge from both historical source and target tasks to improve the predictive performance on the newest target task. Following our generalization error bounds in Corollary~\ref{col:marginal}, we have the following objective function for learning the predictive function of $\mathcal{D}_{t+1}^t$ on the $(N+1)^{\text{th}}$ time stamp.
\begin{equation}\label{eq:framework}
    \begin{aligned}
    &\min_{\theta} J(\theta) = \sum_{j=1}^N \left(\hat{\epsilon}_j^s(\theta) + \hat{\epsilon}_j^t(\theta) \right) + \gamma\cdot \Big(\hat{d}(\mathcal{D}^s_1, \mathcal{D}^t_1; \theta) \\
    &\quad\quad\quad + \sum_{j=1}^{N-1} \hat{d}(\mathcal{D}^s_j, \mathcal{D}^s_{j+1}; \theta) + \sum_{j=1}^N \hat{d}(\mathcal{D}^t_j, \mathcal{D}^t_{j+1}; \theta) \Big)
    \end{aligned}
\end{equation}
where $\hat{d}(\cdot, \cdot; \theta)$ is the empirical distribution discrepancy estimated from finite samples. It can be instantiated with any existing domain discrepancy measures discussed in Corollary~\ref{col:marginal}. We would like to point out that the error bound of Corollary~\ref{col:marginal} is derived in terms of the maximum distribution discrepancy across tasks and across time stamps. But there is no prior knowledge regarding the time stamp with the maximum distribution discrepancy. Therefore, in our framework, we propose to minimize all the distribution discrepancies across tasks and across time stamps.

However, the learned model on the newest target task $\mathcal{D}_{N+1}^t$ might have the issue of catastrophic forgetting such that it performs badly on historical target tasks when updating the new target task. To solve this problem, we would like to learn optimal prior model initialization shared across all the target tasks such that this model can be efficiently fine-tuned on both new and historical target tasks with just a few updates. Figure~\ref{fig:framework} illustrates our proposed \model{} framework with three crucial components: meta-pairs of tasks, meta-training, and meta-testing.

\subsubsection{Meta-Pairs of Tasks}
With the assumption that the source and target tasks are continuously evolving over time, the framework of Eq. (\ref{eq:framework}) is equivalent to sequentially optimizing with respect to the adjacent tasks using standard transfer learning of Eq. (\ref{eq:base_model}).
\begin{equation}\label{eq:new_framework}
    \begin{aligned}
    \min_{\theta} J(\theta) &= \sum_{j=1}^{N-1} \left( \hat{\epsilon}_{j+1}^s(\theta) + \gamma\cdot \hat{d}(\mathcal{D}^s_j, \mathcal{D}^s_{j+1}; \theta) \right)\\
    &\quad + \left(\hat{\epsilon}_1^s(\theta) + \gamma\cdot \hat{d}(\mathcal{D}^s_1, \mathcal{D}^t_1; \theta)\right) \\
    &\quad + \sum_{j=1}^{N} \left( \hat{\epsilon}_j^t(\theta) + \gamma\cdot \hat{d}\left(\mathcal{D}^t_j, \mathcal{D}^t_{j+1}; \theta \right) \right)
    \end{aligned}
\end{equation}
Thus we would like to simply split all the tasks into a set of meta-pairs consisting of two consecutive tasks as shown in Figure~\ref{fig:framework}, and learn the prior model initialization with those meta-pairs of tasks. Different from previous works~\cite{hoffman2014continuous,liu2020learning} with only static source task, we argue that the evolution pattern of the source task can also help improve the performance of \model\ on the newest target task (see more empirical analysis in our experiments). 

\subsubsection{Meta-Training} 
Let $\zeta_j(\theta) = \hat{\epsilon}_{j}^t(\theta) + \gamma\cdot \hat{d}(\mathcal{D}^t_j, \mathcal{D}^t_{j+1}; \theta)$, $\zeta_0(\theta) = \hat{\epsilon}_1^s(\theta) + \gamma\cdot \hat{d}(\mathcal{D}^s_1, \mathcal{D}^t_1; \theta)$ and $\zeta_{-j}(\theta) = \hat{\epsilon}_{j+1}^s(\theta) + \gamma\cdot \hat{d}(\mathcal{D}^s_j, \mathcal{D}^s_{j+1}; \theta)$. Then Eq. (\ref{eq:new_framework}) has a simplified expression $\min_{\theta} J(\theta) = \sum_{k=1-N}^N \zeta_k(\theta)$,
where $\zeta_k$ denotes the objective function of standard transfer learning across tasks and across time stamps. Moreover, it can be formulated as a meta-learning problem~\cite{finn2017model}. That is, the optimal model initialization is learned from historical source and target tasks such that it can be adapted to the newest target task with a few updates. To be more specific, we randomly split the training data from every historical source or target task into one training set $\mathcal{D}_k^{tr}$ and one validation set $\mathcal{D}_k^{val}$. Let $\zeta_k^{tr}$ ($\zeta_k^{val}$) be the loss function of $\zeta_k$ on the training (validation) set. The model initialization $\Tilde{\theta}_N^*$ can be learned as follows.

\begin{equation}\label{eq:meta-train}
    \begin{aligned}
        \Tilde{\theta}_N^* &\leftarrow \arg\min_{\theta} \sum_{k=1-N}^{N-1} \zeta_k^{val}(M_k(\theta)) \\
        M_k(\theta) &\leftarrow \theta - \alpha\cdot \nabla_{\theta} \zeta_k^{tr}(\theta)
    \end{aligned}
\end{equation}
where $M_k: \theta \rightarrow \theta_k$ is a mapping function which learns the optimal task-specific parameter $\theta_k$ from model initialization $\theta$. Following the model-agnostic meta-learning (MAML)~\cite{finn2017model}, $M_k(\theta)$ can be instantiated by one or a few gradient descent updates. Here $\alpha\geq 0$ is the learning rate of the inner loop when training on a specific task. In addition, the training examples of historical target tasks are not labeled. Thus, we propose to sequentially learn the predictive function for every historical target task and generate the pseudo-labels of unlabeled examples as follows.
\begin{equation}\label{eq:inner_update}
\begin{aligned}
    \Tilde{\theta}_{j-1}^* & \leftarrow \arg\min_{\theta} \sum_{k=1-N}^{j-2} \zeta_k^{val}(M_k(\theta)) \\
    \hat{y}_{ij}^{t} & \leftarrow p\left(y|\mathbf{x}_{ij}^{t}; M_{j-1}(\Tilde{\theta}_{j-1}^*)\right)
\end{aligned}
\end{equation}
where $\hat{y}_{ij}^{t}$ is the predicted pseudo-label of input example $\mathbf{x}_{ij}^t$ from the historical target task $\mathcal{D}_j^t$ ($j=1,\cdots,N$). Notice that the training examples with incorrect pseudo-labels might lead to the accumulation of misclassification errors on the new target tasks over time. To mitigate this issue, we propose to select those examples with high prediction confidence. Specifically, we estimate the entropy of the predicted class probability of target examples, and then choose the top $p\%$ with the lowest entropy values. We empirically evaluate this sampling strategy in the experiments.

\subsubsection{Meta-Testing} 
The optimal parameters $\theta_{N+1}$ on the newest target task $\mathcal{D}^t_{N+1}$ could be obtained as follows.
\begin{align}\label{eq:meta-test}
    \theta_{N+1} = M_{N}(\Tilde{\theta}_N^*) \leftarrow \Tilde{\theta}_N^* - \alpha\cdot \nabla_{\Tilde{\theta}_N^*} \zeta_{N}\left( \Tilde{\theta}_N^* \right)
\end{align}
where $\Tilde{\theta}_N^*$ is the optimized model initialization learned in the meta-training phase.

The intuition of this meta-learning framework \model\ can be illustrated as follows. The evolution of dynamic source and target tasks can be represented as a sequential knowledge transfer process across time stamps. But from the perspective of transfer learning~\cite{pan2009survey}, it would be an asymmetric knowledge transfer process for every time stamp, with the goal of maximizing the prediction performance on the new task. That explains why continuous knowledge transfer~\cite{bobu2018adapting,liu2020learning,kumar2020understanding} might have the issue of catastrophic forgetting. However, the continuous evolution of source and target tasks indicates that there might exist some common knowledge transferred across all time stamps. For instance, no matter how the fashion of clothes images changes over time, it follows the basic designs (e.g., shape) for different types of clothes. This common knowledge can be captured by the prior model initialization in our meta-learning framework. It then enables the fast adaptation on the newest target task with only a few model updates.

\section{Experiments}\label{sec:experiments}
\subsection{Experiment Setup}
\subsubsection{Data Sets} We used three publicly available image data sets: Office-31 (with 3 tasks: Amazon, Webcam and DSLR), Image-CLEF (with 4 tasks: B, C, I and P) and Caltran. For Office-31 and Image-CLEF, there are 5 time stamps in the source task and 6 time stamp in the target task (see Appendix for more details). Caltran contains the real-time images captured by a camera at an intersection for two weeks.

\subsubsection{Baseline Methods}
The comparison baselines are given below: (1) static adaptation: SourceOnly that trains only on the source task, DANN~\cite{ganin2016domain}, and MDD~\cite{zhang2019bridging}; (2) multi-source adaptation: MDAN~\cite{zhao2018adversarial}, M3SDA~\cite{peng2019moment}, and DARN~\cite{wen2020domain}; (3) continuous adaptation: CUA~\cite{bobu2018adapting}, TransLATE~\cite{wu2020continuous}, GST~\cite{kumar2020understanding}, and our \model\ with JS-divergence. Here, we merge all the labeled source data into a large one, and then transfer its knowledge to the newest target task for static adaptation methods. For fair comparison, all the multi-source and continuous adaptation methods used both historical source and target data for knowledge transfer, and the target selection strategy for choosing high-quality target examples with pseudo-labels.

\subsubsection{Configuration} We adopted the ResNet-18~\cite{he2016deep} pretrained on ImageNet as the base network for feature extraction, and set $\gamma=0.1$ and $p=80$ for all the experiments\footnote{\url{https://github.com/jwu4sml/L2E}}.

\begin{figure}[!t]
\centering
\subfigure[Webcam $\rightarrow$ DSLR]{\label{fig:b}\includegraphics[width=4.1cm]{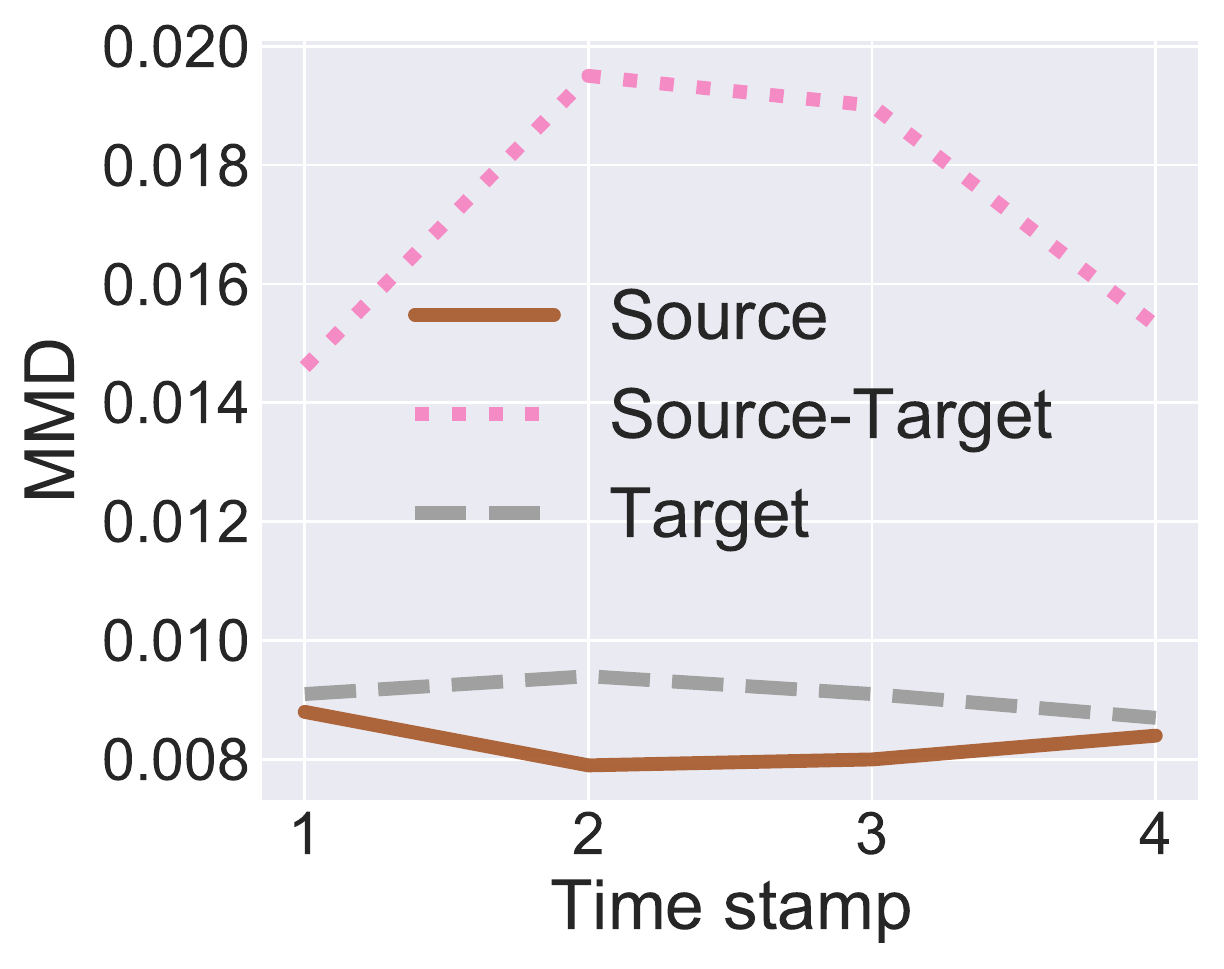}}
\subfigure[B $\rightarrow$ P]{\label{fig:a}\includegraphics[width=4.1cm]{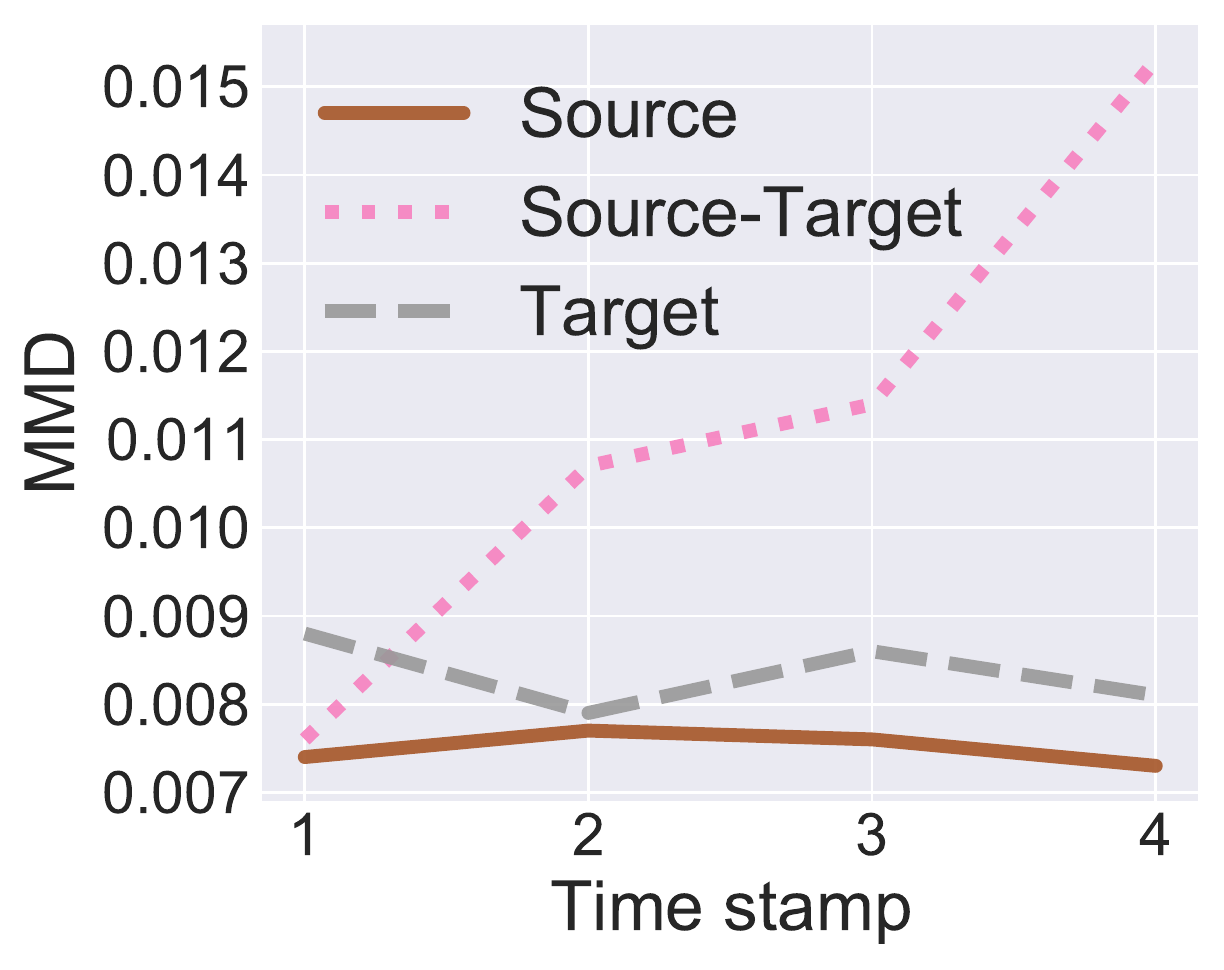}}
\caption{Task evolution on Office-31 (Webcam $\rightarrow$ DSLR) and Image-CLEF (B $\rightarrow$ P) where ``Source": $\hat{d}(\mathcal{D}^s_j, \mathcal{D}^s_{j+1})$, ``Source-Target": $\hat{d}(\mathcal{D}^s_j, \mathcal{D}^t_{j})$, and ``Target": $\hat{d}(\mathcal{D}^t_j, \mathcal{D}^t_{j+1})$}\label{fig:MMD}
\end{figure}

\begin{table}[!t]
    \centering
    \small
    \begin{tabular}{l|cc|cc}
    \toprule
        \multirow{2}{*}{Method} & \multicolumn{2}{c|}{Amazon $\rightarrow$ Webcam} & \multicolumn{2}{c}{Webcam $\rightarrow$ DSLR} \\\cline{2-5}
         & Acc & H-Acc & Acc & H-Acc \\\midrule
        SourceOnly & 0.17$_{\pm \text{0.02}}$ & 0.35$_{\pm \text{0.02}}$ & 0.78$_{\pm \text{0.02}}$ & 0.88$_{\pm \text{0.03}}$\\
        DANN & 0.32$_{\pm \text{0.03}}$ & 0.45$_{\pm \text{0.04}}$ & 0.84$_{\pm \text{0.01}}$ & 0.90$_{\pm \text{0.01}}$\\
        MDD & 0.32$_{\pm \text{0.01}}$ & 0.45$_{\pm \text{0.01}}$ & 0.85$_{\pm \text{0.02}}$ & 0.91$_{\pm \text{0.01}}$\\\midrule
        MDAN & 0.49$_{\pm \text{0.03}}$ & 0.57$_{\pm \text{0.01}}$ & 0.88$_{\pm \text{0.00}}$ & 0.92$_{\pm \text{0.01}}$\\
        M3SDA & 0.49$_{\pm \text{0.03}}$ & 0.57$_{\pm \text{0.03}}$ & 0.81$_{\pm \text{0.04}}$ & 0.86$_{\pm \text{0.03}}$\\
        DARN & 0.45$_{\pm \text{0.02}}$ & 0.49$_{\pm \text{0.03}}$ & 0.69$_{\pm \text{0.03}}$ & 0.74$_{\pm \text{0.03}}$\\\midrule
        CUA & 0.48$_{\pm \text{0.01}}$ &\bf 0.58$_{\pm \text{0.01}}$ & 0.85$_{\pm \text{0.03}}$ & 0.89$_{\pm \text{0.04}}$\\
        TransLATE & 0.50$_{\pm \text{0.01}}$ &\bf 0.58$_{\pm \text{0.02}}$ & 0.86$_{\pm \text{0.02}}$ & 0.90$_{\pm \text{0.02}}$\\
        GST & 0.43$_{\pm \text{0.01}}$ & 0.45$_{\pm \text{0.00}}$ & 0.84$_{\pm \text{0.02}}$ & 0.80$_{\pm \text{0.01}}$ \\\midrule
        \model{} (ours) &\bf 0.52$_{\pm \text{0.02}}$ &\bf 0.58$_{\pm \text{0.01}}$ &\bf 0.89$_{\pm \text{0.01}}$ &\bf 0.95$_{\pm \text{0.00}}$ \\
    \bottomrule
    \end{tabular}
    \caption{Transfer learning accuracy on Office-31}
    \label{exp:office31}
\end{table}

\begin{table}[!t]
    \centering
    \small
    \begin{tabular}{l|cc|cc}
    \toprule
        \multirow{2}{*}{Method} & \multicolumn{2}{c|}{I $\rightarrow$ C}                       & \multicolumn{2}{c}{B $\rightarrow$ P}  \\\cline{2-5}
         & Acc & H-Acc & Acc & H-Acc \\\midrule
        SourceOnly & 0.26$_{\pm \text{0.01}}$ & 0.51$_{\pm \text{0.01}}$ & 0.24$_{\pm \text{0.00}}$ & 0.43$_{\pm \text{0.01}}$ \\
        DANN & 0.36$_{\pm \text{0.00}}$ & 0.58$_{\pm \text{0.01}}$ & 0.27$_{\pm \text{0.01}}$ & 0.43$_{\pm \text{0.00}}$ \\
        MDD & 0.41$_{\pm \text{0.01}}$ & 0.62$_{\pm \text{0.01}}$ & 0.28$_{\pm \text{0.03}}$ & 0.42$_{\pm \text{0.01}}$ \\\midrule
        MDAN & 0.62$_{\pm \text{0.03}}$ & 0.77$_{\pm \text{0.00}}$ & 0.37$_{\pm \text{0.05}}$ & 0.51$_{\pm \text{0.02}}$ \\
        M3SDA & 0.56$_{\pm \text{0.03}}$ & 0.74$_{\pm \text{0.01}}$ & 0.39$_{\pm \text{0.02}}$ & 0.52$_{\pm \text{0.02}}$ \\
        DARN & 0.55$_{\pm \text{0.02}}$ & 0.76$_{\pm \text{0.02}}$ & 0.39$_{\pm \text{0.02}}$ & 0.52$_{\pm \text{0.01}}$ \\\midrule
        CUA & 0.58$_{\pm \text{0.01}}$ & 0.74$_{\pm \text{0.01}}$ & 0.36$_{\pm \text{0.03}}$ & 0.51$_{\pm \text{0.00}}$ \\
        TransLATE & 0.64$_{\pm \text{0.01}}$ & 0.76$_{\pm \text{0.00}}$ & 0.40$_{\pm \text{0.03}}$ & 0.55$_{\pm \text{0.01}}$ \\
        GST & 0.39$_{\pm \text{0.01}}$ & 0.54$_{\pm \text{0.03}}$ & 0.32$_{\pm \text{0.01}}$ & 0.31$_{\pm \text{0.02}}$ \\\midrule
        \model{} (ours) &\bf 0.66$_{\pm \text{0.02}}$ &\bf 0.80$_{\pm \text{0.01}}$ &\bf 0.44$_{\pm \text{0.04}}$ &\bf 0.57$_{\pm \text{0.02}}$ \\
    \bottomrule
    \end{tabular}
    \caption{Transfer learning accuracy on Image-CLEF}
    \label{exp:image_clef}
\end{table}

\begin{figure}[!t]
\centering
\subfigure[Acc]{\label{fig:a}\includegraphics[width=4.23cm]{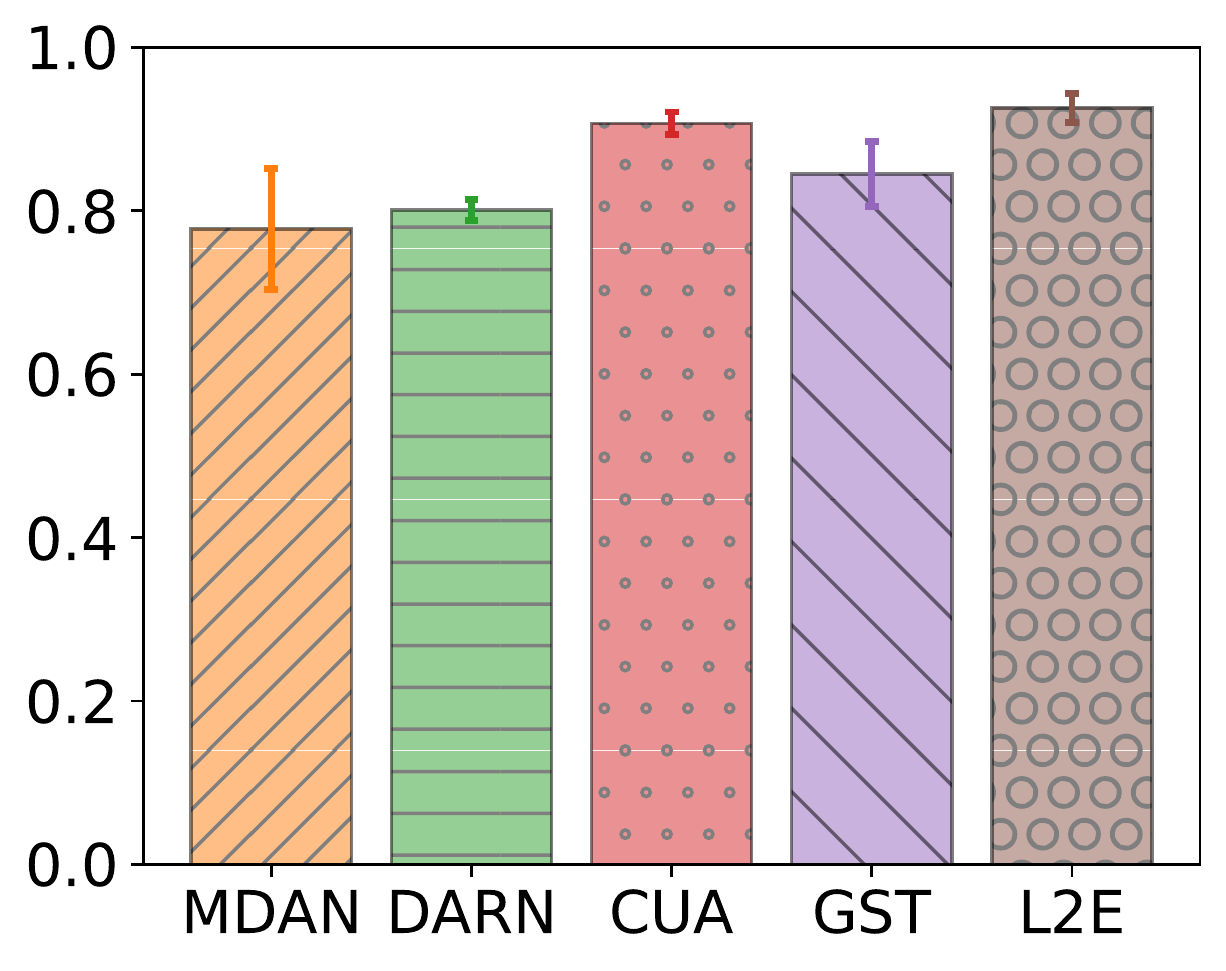}}
\subfigure[H-Acc]{\label{fig:b}\includegraphics[width=4.23cm]{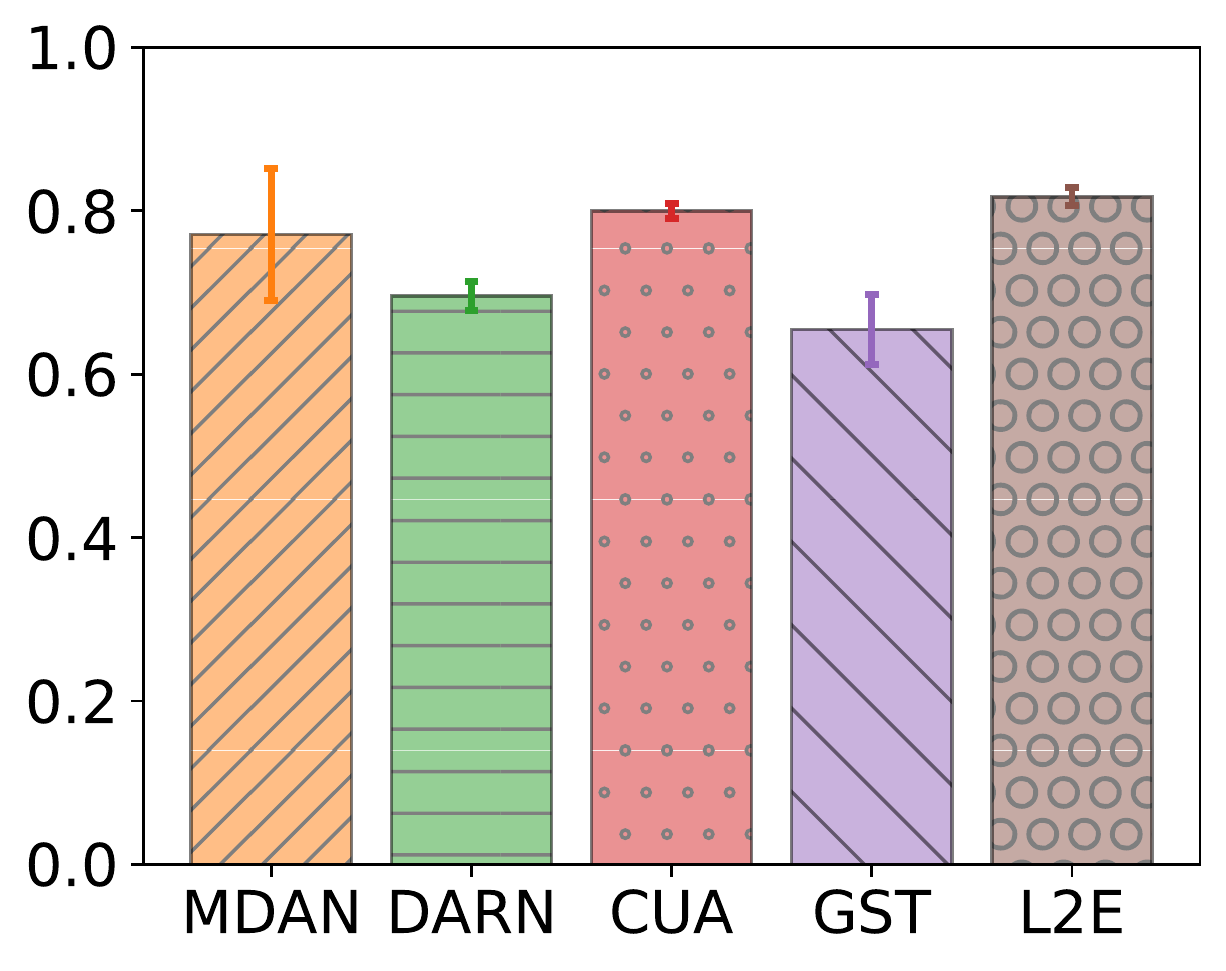}}
\caption{Transfer learning accuracy on Caltran}\label{fig:Caltran}
\end{figure}

\subsection{Results}
Figure~\ref{fig:MMD} shows the results of domain discrepancy via MMD, including (i) the evolution of source task, i.e., $\hat{d}(\mathcal{D}^s_j, \mathcal{D}^s_{j+1})$, (ii) the evolution of target task, i.e., $\hat{d}(\mathcal{D}^t_j, \mathcal{D}^t_{j+1})$, and (iii) the evolution of task relatedness, i.e., $\hat{d}(\mathcal{D}^s_j, \mathcal{D}^t_{j})$. It indicates that in both Office-31 and Image-CLEF, the source and target tasks are changing smoothly, whereas the relatedness between source and target tasks is decreasing over time. 
Table~\ref{exp:office31}, Table~\ref{exp:image_clef} and Figure~\ref{fig:Caltran} provide the transfer learning results on the dynamic tasks where the classification accuracies on the newest target task (Acc) and all the historical target tasks (H-Acc) are reported. We run all the experiments five times and report the mean and standard deviation of classification accuracies (the best results are indicated in bold). It can be observed that: (1) compared to static methods, the multi-source and continuous adaptation methods can achieve much better classification performance by leveraging the historical knowledge; (2) our proposed framework \model{} outperforms state-of-the-art baselines in both the newest target task and all the historical target tasks. This confirms that \model\ mitigates the issue of catastrophic forgetting on historical tasks when learning the new task.

\begin{table}[!t]
    \centering
    \small
    \begin{tabular}{l|cc}
    \toprule
        Method & Acc & H-Acc  \\\midrule
        \model\ w/o source evolution & 0.38$_{\pm \text{0.04}}$ & 0.57$_{\pm \text{0.02}}$ \\
        \model\ w merged source & 0.35$_{\pm \text{0.07}}$ & 0.51$_{\pm \text{0.02}}$ \\
        \model\ w/o historical target & 0.21$_{\pm \text{0.0}}$ & 0.46$_{\pm \text{0.02}}$ \\
        \model\ w all pairs & 0.39$_{\pm \text{0.03}}$ & 0.57$_{\pm \text{0.010}}$ \\
        \model & \bf 0.44$_{\pm \text{0.04}}$ &\bf 0.57$_{\pm \text{0.02}}$ \\
    \bottomrule
    \end{tabular}
    \caption{Ablation study on Image-CLEF (B $\rightarrow$ P)}
    \label{tab:ablation_study}
\end{table}


\begin{figure}[!t]
  \centering
  \begin{minipage}[b]{.23\textwidth}
    \includegraphics[width=\textwidth]{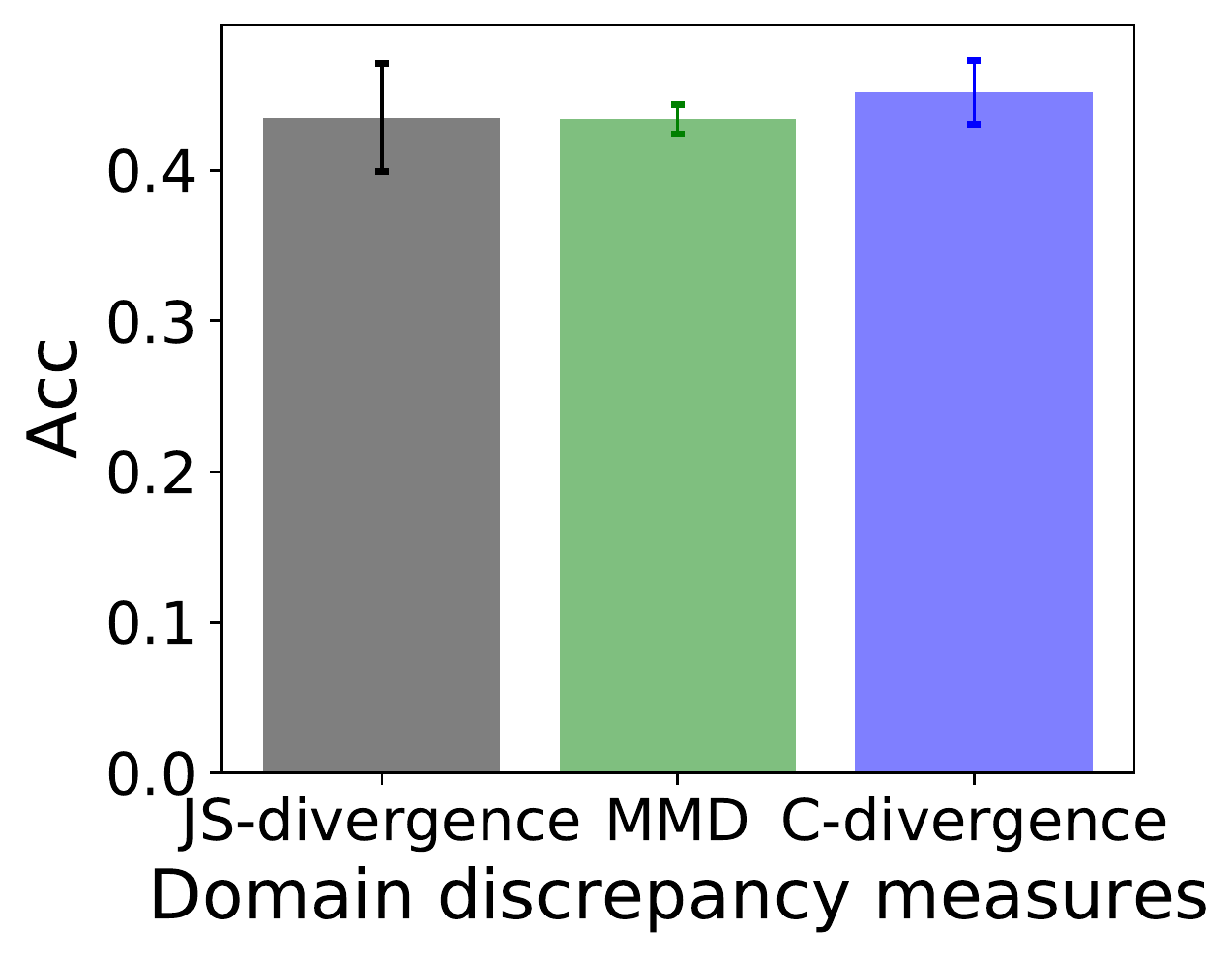}
    \caption{Impact of domain discrepancy measures}\label{fig:domain_discrepancy}
  \end{minipage}
  \hfill
  \begin{minipage}[b]{.23\textwidth}
    \includegraphics[width=\textwidth]{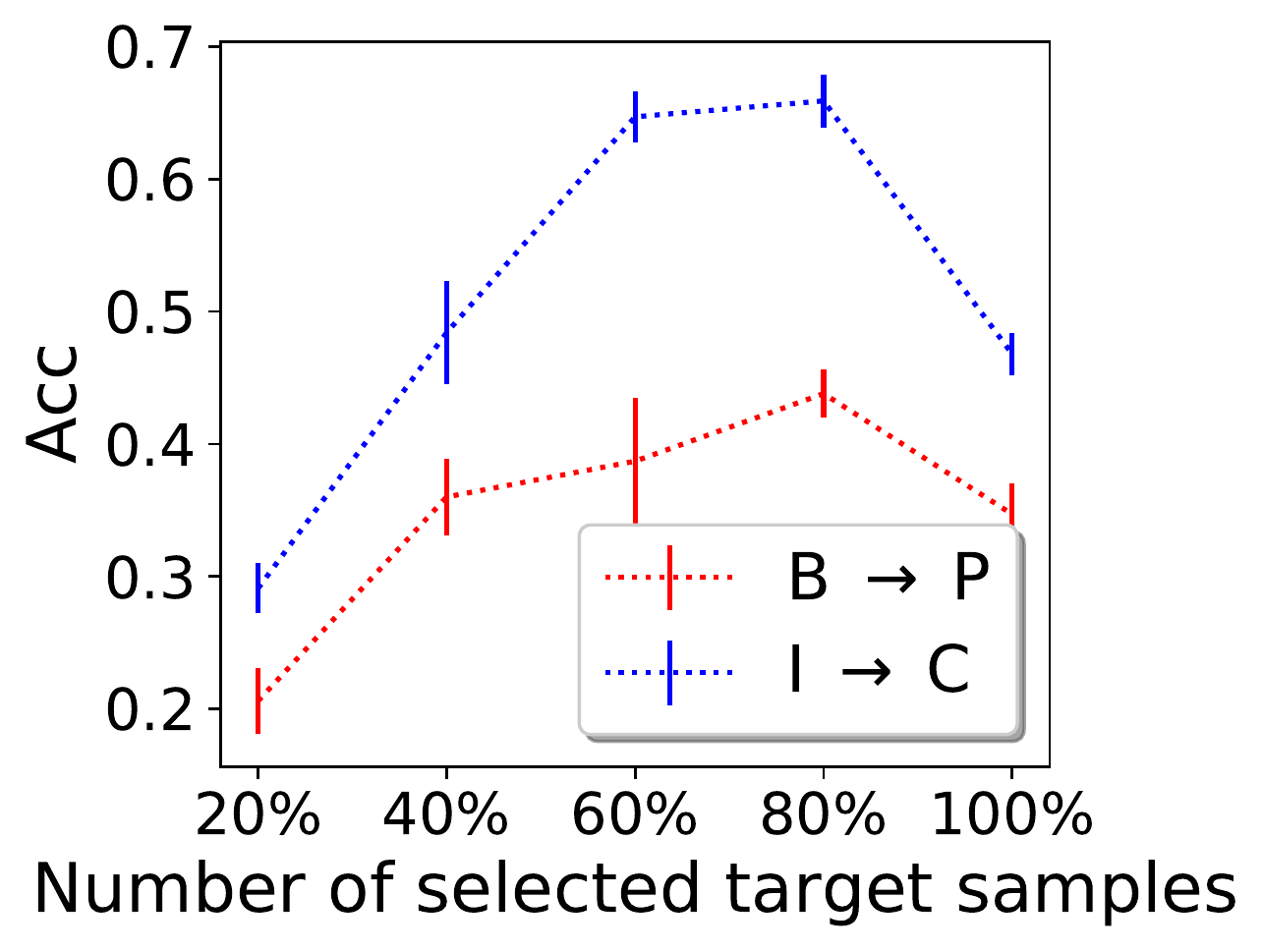}
    \caption{Impact of sampling selection strategy}\label{fig:sampling_selection}
  \end{minipage}
\end{figure}

\subsection{Case Studies}
Table~\ref{tab:ablation_study} reports the results of several variants on Image-CLEF (B $\rightarrow$ P): (i) \model\ w/o source evolution: using only source data at the initial time stamp; (ii) \model\ w merged source: merging all source data into a large one; (iii) \model\ w/o historical target: transferring the dynamic source tasks to the newest target task directly without historical target knowledge. It is observed that the evolution knowledge from historical source and target tasks could indeed improve the performance of \model\ in dynamic transfer learning. Besides, we also consider to generate the meta-pairs from any two historical tasks (indicated in ``\model\ w all pairs" in Table~\ref{tab:ablation_study}). It could not outperform \model\ with only the meta-pairs from consecutive tasks. One explanation is that it might generate the unrelated meta-pairs of tasks.
Figure~\ref{fig:domain_discrepancy} shows the results of \model\ instantiated with JS-divergence~\cite{ganin2016domain,pmlr-v139-acuna21a}, MMD~\cite{long2015learning} and $\mathcal{C}$-divergence~\cite{wu2020continuous} on Image-CLEF (B $\rightarrow$ P). It indicates that our \model\ framework is flexible to incorporate any domain discrepancy measures. Figure~\ref{fig:sampling_selection} shows the impact of sampling selection ratio $p\%$ on \model\ where the classification accuracies on the newest target task (Acc) are reported on Image-CLEF. It confirms that selecting the historical target examples with high confidence positively affects the performance of \model. Thus we choose $p=80$ for our experiments.

\section{Conclusion}\label{sec:conclusion}
In this paper, we study the transfer learning problem with dynamic source and target tasks. We show the error bounds of dynamic transfer learning on the newest target task in terms of historical source and target task knowledge. Then we propose a generic meta-learning framework \model{} by minimizing the error upper bounds. Empirical results demonstrate the effectiveness of the proposed \model{} framework.

\section*{Acknowledgements}
This work is supported by National Science Foundation under Award No. IIS-1947203, IIS-2117902, IIS-2137468, and Agriculture and Food Research Initiative (AFRI) grant no. 2020-67021-32799/project accession no.1024178 from the USDA National Institute of Food and Agriculture. The views and conclusions are those of the authors and should not be interpreted as representing the official policies of the funding agencies or the government.

\bibliographystyle{named}
\bibliography{ijcai22}

\begin{thebibliography}{}

\bibitem[\protect\citeauthoryear{Acuna \bgroup \em et al.\egroup
  }{2021}]{pmlr-v139-acuna21a}
David Acuna, Guojun Zhang, Marc~T. Law, and Sanja Fidler.
\newblock $f$-domain adversarial learning: Theory and algorithms.
\newblock In {\em ICML}, 2021.

\bibitem[\protect\citeauthoryear{Al-Halah \bgroup \em et al.\egroup
  }{2017}]{al2017fashion}
Ziad Al-Halah, Rainer Stiefelhagen, and Kristen Grauman.
\newblock Fashion forward: Forecasting visual style in fashion.
\newblock In {\em ICCV}, 2017.

\bibitem[\protect\citeauthoryear{Ben-David \bgroup \em et al.\egroup
  }{2010}]{ben2010theory}
Shai Ben-David, John Blitzer, Koby Crammer, Alex Kulesza, Fernando Pereira, and
  Jennifer~Wortman Vaughan.
\newblock A theory of learning from different domains.
\newblock {\em Machine learning}, 2010.

\bibitem[\protect\citeauthoryear{Bobu \bgroup \em et al.\egroup
  }{2018}]{bobu2018adapting}
Andreea Bobu, Eric Tzeng, Judy Hoffman, and Trevor Darrell.
\newblock Adapting to continuously shifting domains.
\newblock In {\em ICLR Workshop}, 2018.

\bibitem[\protect\citeauthoryear{Fallah \bgroup \em et al.\egroup
  }{2020}]{fallah2020convergence}
Alireza Fallah, Aryan Mokhtari, and Asuman Ozdaglar.
\newblock On the convergence theory of gradient-based model-agnostic
  meta-learning algorithms.
\newblock In {\em AISTATS}, 2020.

\bibitem[\protect\citeauthoryear{Finn \bgroup \em et al.\egroup
  }{2017}]{finn2017model}
Chelsea Finn, Pieter Abbeel, and Sergey Levine.
\newblock Model-agnostic meta-learning for fast adaptation of deep networks.
\newblock In {\em ICML}, 2017.

\bibitem[\protect\citeauthoryear{Finn \bgroup \em et al.\egroup
  }{2019}]{finn2019online}
Chelsea Finn, Aravind Rajeswaran, Sham Kakade, and Sergey Levine.
\newblock Online meta-learning.
\newblock In {\em ICML}, 2019.

\bibitem[\protect\citeauthoryear{Ganin \bgroup \em et al.\egroup
  }{2016}]{ganin2016domain}
Yaroslav Ganin, Evgeniya Ustinova, Hana Ajakan, Pascal Germain, Hugo
  Larochelle, Fran{\c{c}}ois Laviolette, Mario Marchand, and Victor Lempitsky.
\newblock Domain-adversarial training of neural networks.
\newblock {\em JMLR}, 17(1):2096--2030, 2016.

\bibitem[\protect\citeauthoryear{Ghifary \bgroup \em et al.\egroup
  }{2016}]{ghifary2016scatter}
Muhammad Ghifary, David Balduzzi, W~Bastiaan Kleijn, and Mengjie Zhang.
\newblock Scatter component analysis: A unified framework for domain adaptation
  and domain generalization.
\newblock {\em TPAMI}, 2016.

\bibitem[\protect\citeauthoryear{Gretton \bgroup \em et al.\egroup
  }{2012}]{gretton2012kernel}
Arthur Gretton, Karsten~M Borgwardt, Malte~J Rasch, Bernhard Sch{\"o}lkopf, and
  Alexander Smola.
\newblock A kernel two-sample test.
\newblock {\em JMLR}, 2012.

\bibitem[\protect\citeauthoryear{He \bgroup \em et al.\egroup
  }{2016}]{he2016deep}
Kaiming He, Xiangyu Zhang, Shaoqing Ren, and Jian Sun.
\newblock Deep residual learning for image recognition.
\newblock In {\em CVPR}, pages 770--778, 2016.

\bibitem[\protect\citeauthoryear{Hoffman \bgroup \em et al.\egroup
  }{2014}]{hoffman2014continuous}
Judy Hoffman, Trevor Darrell, and Kate Saenko.
\newblock Continuous manifold based adaptation for evolving visual domains.
\newblock In {\em CVPR}, 2014.

\bibitem[\protect\citeauthoryear{Kumar \bgroup \em et al.\egroup
  }{2020}]{kumar2020understanding}
Ananya Kumar, Tengyu Ma, and Percy Liang.
\newblock Understanding self-training for gradual domain adaptation.
\newblock In {\em ICML}, 2020.

\bibitem[\protect\citeauthoryear{Li and Hoiem}{2017}]{li2017learning}
Zhizhong Li and Derek Hoiem.
\newblock Learning without forgetting.
\newblock {\em TPAMI}, 2017.

\bibitem[\protect\citeauthoryear{Liu \bgroup \em et al.\egroup
  }{2020}]{liu2020learning}
Hong Liu, Mingsheng Long, Jianmin Wang, and Yu~Wang.
\newblock Learning to adapt to evolving domains.
\newblock {\em NeurIPS}, 2020.

\bibitem[\protect\citeauthoryear{Long \bgroup \em et al.\egroup
  }{2015}]{long2015learning}
Mingsheng Long, Yue Cao, Jianmin Wang, and Michael Jordan.
\newblock Learning transferable features with deep adaptation networks.
\newblock In {\em ICML}, 2015.

\bibitem[\protect\citeauthoryear{Mansour \bgroup \em et al.\egroup
  }{2009}]{mansour2009domain}
Yishay Mansour, Mehryar Mohri, and Afshin Rostamizadeh.
\newblock Domain adaptation: Learning bounds and algorithms.
\newblock In {\em COLT}, 2009.

\bibitem[\protect\citeauthoryear{Mohri and Medina}{2012}]{mohri2012new}
Mehryar Mohri and Andres~Munoz Medina.
\newblock New analysis and algorithm for learning with drifting distributions.
\newblock In {\em ALT}, 2012.

\bibitem[\protect\citeauthoryear{Pan and Yang}{2009}]{pan2009survey}
Sinno~Jialin Pan and Qiang Yang.
\newblock A survey on transfer learning.
\newblock {\em TKDE}, 2009.

\bibitem[\protect\citeauthoryear{Peng \bgroup \em et al.\egroup
  }{2019}]{peng2019moment}
Xingchao Peng, Qinxun Bai, Xide Xia, Zijun Huang, Kate Saenko, and Bo~Wang.
\newblock Moment matching for multi-source domain adaptation.
\newblock In {\em ICCV}, 2019.

\bibitem[\protect\citeauthoryear{Rafailidis and
  Nanopoulos}{2015}]{rafailidis2015modeling}
Dimitrios Rafailidis and Alexandros Nanopoulos.
\newblock Modeling users preference dynamics and side information in
  recommender systems.
\newblock {\em IEEE Transactions on Systems, Man, and Cybernetics: Systems},
  46(6):782--792, 2015.

\bibitem[\protect\citeauthoryear{Rosenstein \bgroup \em et al.\egroup
  }{2005}]{rosenstein2005transfer}
Michael~T Rosenstein, Zvika Marx, Leslie~Pack Kaelbling, and Thomas~G
  Dietterich.
\newblock To transfer or not to transfer.
\newblock In {\em NIPS workshop}, 2005.

\bibitem[\protect\citeauthoryear{Sun \bgroup \em et al.\egroup
  }{2019}]{sun2019meta}
Qianru Sun, Yaoyao Liu, Tat-Seng Chua, and Bernt Schiele.
\newblock Meta-transfer learning for few-shot learning.
\newblock In {\em CVPR}, 2019.

\bibitem[\protect\citeauthoryear{Wang \bgroup \em et al.\egroup
  }{2020}]{wang2020continuously}
Hao Wang, Hao He, and Dina Katabi.
\newblock Continuously indexed domain adaptation.
\newblock In {\em ICML}, 2020.

\bibitem[\protect\citeauthoryear{Wen \bgroup \em et al.\egroup
  }{2020}]{wen2020domain}
Junfeng Wen, Russell Greiner, and Dale Schuurmans.
\newblock Domain aggregation networks for multi-source domain adaptation.
\newblock In {\em ICML}, 2020.

\bibitem[\protect\citeauthoryear{Wu and He}{2020}]{wu2020continuous}
Jun Wu and Jingrui He.
\newblock Continuous transfer learning with label-informed distribution
  alignment.
\newblock {\em arXiv preprint arXiv:2006.03230}, 2020.

\bibitem[\protect\citeauthoryear{Wu and He}{2021}]{wu2021indirect}
Jun Wu and Jingrui He.
\newblock Indirect invisible poisoning attacks on domain adaptation.
\newblock In {\em KDD}, 2021.

\bibitem[\protect\citeauthoryear{Zhang \bgroup \em et al.\egroup
  }{2019}]{zhang2019bridging}
Yuchen Zhang, Tianle Liu, Mingsheng Long, and Michael Jordan.
\newblock Bridging theory and algorithm for domain adaptation.
\newblock In {\em ICML}, 2019.

\bibitem[\protect\citeauthoryear{Zhao \bgroup \em et al.\egroup
  }{2018}]{zhao2018adversarial}
Han Zhao, Shanghang Zhang, Guanhang Wu, Jos{\'e}~MF Moura, Joao~P Costeira, and
  Geoffrey~J Gordon.
\newblock Adversarial multiple source domain adaptation.
\newblock {\em NeurIPS}, 2018.

\bibitem[\protect\citeauthoryear{Zhou \bgroup \em et al.\egroup
  }{2017}]{Multic2_Yao_2017}
Yao Zhou, Lei Ying, and Jingrui He.
\newblock Multi{C}\({}^{2}\): an optimization framework for learning from task
  and worker dual heterogeneity.
\newblock In {\em SDM}, 2017.

\bibitem[\protect\citeauthoryear{Zhou \bgroup \em et al.\egroup
  }{2019}]{zhou2019multi}
Yao Zhou, Lei Ying, and Jingrui He.
\newblock Multi-task crowdsourcing via an optimization framework.
\newblock {\em TKDD}, 13(3):1--26, 2019.

\end{thebibliography}

\clearpage
\onecolumn
\appendix
\section{Appendices}

\begin{definition} ({\bf Rademacher Complexity})
Let $\mathcal{H}$ be a set of real-valued functions mapping from $\mathcal{X}$ to $\mathbb{R}$. Given a sample $S\in \mathcal{X}^{m}$, the empirical Rademacher complexity of $\mathcal{H}$ for the sample $S$ is defined as follows.
\begin{align*}
    \hat{\Re}_{S}(\mathcal{H}) = \mathbb{E}_{\bm\sigma}\left[\sup_{h\in \mathcal{H}} \frac{2}{m} \sum_{i=1}^m \sigma_i h(\mathbf{x}_i) \Big| S = (\mathbf{x}_1, \cdots, \mathbf{x}_m) \right]
\end{align*}
where $\bm\sigma = (\sigma_1,\cdots,\sigma_m)$ is a vector of $m$ independent uniform random variables taking values in $\{-1, +1\}$. The Rademacher complexity of $\mathcal{H}$ is defined as the expectation of $\hat{\Re}_{S}(\mathcal{H})$ over all samples $S=(\mathbf{x}_1, \cdots, \mathbf{x}_m)$ of size $m$:
\begin{align*}
    \Re_S(\mathcal{H}) = \mathbb{E}_{m}\left[ \hat{\Re}_{S}(\mathcal{H}) \Big| |S|=m  \right]
\end{align*}
\end{definition}

\begin{lemma}\label{McDiarmid} ({\bf\em McDiarmid's inequality})
Let $X_1, \cdots, X_m$ be independently random variables taking values in the set $\mathcal{X}$ and $f: \mathcal{X}^m \rightarrow \mathbb{R}$ be a function over $X_1, \cdots, X_m$ that satisfies $\forall i, \forall x_1, \cdots, x_m, x'_i \in \mathcal{X}$,
\begin{equation*}
    \left| f(x_1,\cdots,x_i,\cdots,x_m) - f(x_1,\cdots,x'_i,\cdots,x_m) \right| \leq c_i
\end{equation*}
Then, for any $\epsilon > 0$,
\begin{equation*}
    \mathrm{Pr}\left[ f-\mathbb{E}[f] \geq \epsilon \right] \leq \exp{\left( \frac{-2\epsilon^2}{\sum_{i=1}^m c_i^2} \right)}
\end{equation*}
\end{lemma}

\subsection{Proof of Theorem~\ref{continuous_transfer_learning_bound}}\label{sec:proof_of_l1bound}
Theorem \ref{continuous_transfer_learning_bound} assumes that the loss function $\mathcal{L}(\cdot, \cdot)$ is $\mu$-admissible and obeys the triangle inequality. Given a class of functions $\mathcal{H}_{\mathcal{L}} = \{(\mathbf{x}, y) \mapsto \mathcal{L}(h(\mathbf{x}), y): h\in \mathcal{H} \}$, for any $\delta>0$ and $h\in \mathcal{H}$, with probability at least $1-\delta$, the expected error $\epsilon_{N+1}^t$ for the newest target task $\mathcal{D}_{N+1}^t$ is bounded by
\begin{align*}
    \epsilon_{{N+1}}^t(h) &\leq \frac{1}{2N} \sum_{j=1}^N \left(\hat{\epsilon}_j^s(h) + \hat{\epsilon}_j^t(h) \right) + \frac{N+2}{2} \left( \Tilde{d} + \Tilde{\lambda} \right) + \Tilde{\Re}(\mathcal{H}_{\mathcal{L}}) + \frac{\mu}{N}\sqrt{\frac{\log{\frac{1}{\delta}}}{2\Tilde{m}}}
\end{align*}
where $\Tilde{d} =\mu\cdot \max\big\{\max_{1\leq j\leq N-1}d_1(\mathcal{D}^s_j, \mathcal{D}^s_{j+1}), d_1(\mathcal{D}^s_1, \mathcal{D}^t_1),$ $\max_{1\leq j\leq N}d_1(\mathcal{D}^t_j, \mathcal{D}^t_{j+1}) \big\}$, $\Tilde{\lambda} =\mu\cdot \max\big\{\max_{1\leq j\leq N-1}$ $ \lambda_*(\mathcal{D}^s_j, \mathcal{D}^s_{j+1}), \lambda_*(\mathcal{D}^s_1, \mathcal{D}^t_1), \max_{1\leq j\leq N}\lambda_*(\mathcal{D}^t_j, \mathcal{D}^t_{j+1}) \big\}$ and $\lambda_*$ measures the difference of the labeling functions across task and across time stamps, i.e., $\lambda_*(\mathcal{D}^s_1, \mathcal{D}^t_1) = \min\{\mathbb{E}_{\mathcal{D}^s_1}[|f_1^s(\mathbf{x}) -f_1^t(\mathbf{x})|], \mathbb{E}_{\mathcal{D}^t_1}[|f_1^s(\mathbf{x}) -f_1^t(\mathbf{x})|]\}$. $\Tilde{\Re}(\mathcal{H}_{\mathcal{L}}) = \frac{1}{2N}\sum_{j=1}^N \big({\Re}_{\mathcal{D}_j^s}(\mathcal{H}_{\mathcal{L}}) + {\Re}_{\mathcal{D}_j^t}(\mathcal{H}_{\mathcal{L}}) \big)$ and $\Tilde{m}=$ $\sum_{j=1}^N(m_j^s+m_j^t)$ is the total number of training examples from source and historical target tasks.
\begin{proof}
Let $\mathcal{D}^s_j(\mathbf{x},y)$ be the source distribution and $\mathcal{D}^t_j(\mathbf{x},y)$ be the target distribution at time stamp $j$. For any sample set
$\mathcal{B}=\left(\{(\mathbf{x}_{i1}^s, y_{i1}^s)\}_{i=1}^{m_{1}^s}, \cdots, \{(\mathbf{x}_{iN}^s, y_{iN}^s)\}_{i=1}^{m^s_N} , \{(\mathbf{x}_{i1}^t, y_{i1}^t)\}_{i=1}^{m_{1}^t}, \cdots, \{(\mathbf{x}_{iN}^t, y_{iN}^t)\}_{i=1}^{m^t_N} \right) \in \left( \mathcal{X}\times \mathcal{Y} \right)^{\Tilde{m}}$ sampled from the product distribution $\mathcal{D}(\mathbf{x},y)= \mathcal{D}^s_1(\mathbf{x},y)^{m_1^s} \otimes \cdots \otimes \mathcal{D}^s_N(\mathbf{x},y)^{m_N^s} \otimes \mathcal{D}^t_1(\mathbf{x},y)^{m_{1}^t} \otimes \cdots \otimes \mathcal{D}^t_N(\mathbf{x},y)^{m_N^t}$, we define a function $g$ over $\mathcal{B}$ as follows.
\begin{equation*}
    g(\mathcal{B}) = \sup_{h\in \mathcal{H}} \epsilon_{{N+1}}^t(h) - \frac{1}{2N} \sum_{j=1}^N \left(\hat{\epsilon}_j^s(h) + \hat{\epsilon}_j^t(h) \right)
\end{equation*}
where $\hat{\epsilon}_{j}^s(h)=\frac{1}{m_{j}^s} \sum_{i=1}^{m_{j}^s} \mathcal{L}\left(h(\mathbf{x}_{ij}^s), y_{ij}^s\right)$ and $\hat{\epsilon}_{j}^t(h)=\frac{1}{m_{j}^t} \sum_{i=1}^{m_{j}^t} \mathcal{L}\left(h(\mathbf{x}_{ij}^t), y_{ij}^t\right)$ for all $j=1,\cdots,N$. Let $\mathcal{B}$ and $\mathcal{B}'$ be two sample sets containing only one different source sample at $(\mathbf{x}_{ij}^s, y_{ij}^s)$ and $(\mathbf{\bar{x}}_{ij}^s, \bar{y}_{ij}^s)$, then we have
\begin{equation*}
    \left| g(\mathcal{B}) - g(\mathcal{B}') \right| \leq \frac{1}{2m_j^sN}\sup_{h\in \mathcal{H}} \left| \mathcal{L}\left(h(\mathbf{x}_{ij}^s), y_{ij}^s \right) - \mathcal{L}\left(h(\mathbf{\bar{x}}_{ij}^s), \bar{y}_{ij}^s\right) \right| \leq \frac{2\mu}{2m_j^sN} \leq \frac{\mu}{\Tilde{m}N}
\end{equation*}
The same result holds for different target sample.
Based on McDiarmid's inequality (see Lemma~\ref{McDiarmid}), we have for any $\epsilon > 0$
\begin{equation*}
    \mathrm{Pr}\left[  g(\mathcal{B}) -\mathbb{E}_{\mathcal{B}} \left[g(\mathcal{B}) \right]  \geq \epsilon \right] \leq \exp{\left( \frac{-2\Tilde{m}N^2\epsilon^2}{\mu^2} \right)}
\end{equation*}
Then, for any $\delta > 0$, with probability at least $1-\delta$, the following holds
\begin{equation*}
    g(\mathcal{B}) \leq \mathbb{E}_{\mathcal{B}} \left[g(\mathcal{B})\right] + \frac{\mu}{N}\sqrt{\frac{\log{\frac{1}{\delta}}}{2\Tilde{m}}}
\end{equation*}

In addition, for any $h\in \mathcal{H}$, we have
\begin{align*}
    \epsilon_1^t(h) &= \epsilon_1^t(h) + \epsilon_1^s(h) - \epsilon_1^s(h) + \epsilon_1^s(h, f_1^t) - \epsilon_1^s(h, f_1^t) \\
    &\leq \epsilon_1^s(h) + \left| \epsilon_1^s(h, f_1^t) - \epsilon_1^s(h, f_1^s) \right| + \left| \epsilon_1^t(h, f_1^t) - \epsilon_1^s(h, f_1^t) \right| \\
    &\leq \epsilon_1^s(h) + \mathbb{E}_{\mathcal{D}_1^s}\left[\left| \mathcal{L}\left(h(\mathbf{x}_{1}^s), f^t_1(\mathbf{x}_{1}^s) \right) - \mathcal{L}\left(h(\mathbf{x}_{1}^s), f^s_1(\mathbf{x}_{1}^s) \right) \right|\right] + \left| \epsilon_1^t(h, f_1^t) - \epsilon_1^s(h, f_1^t) \right| \\
    &\leq \epsilon_1^s(h) + \mu \mathbb{E}_{\mathcal{D}_1^s}\left[\left| f^t_1(\mathbf{x}_{1}^s) - f^s_1(\mathbf{x}_{1}^s) \right|\right] + \int \left|\mathcal{D}_1^{t}(\mathbf{x}) - \mathcal{D}_1^s(\mathbf{x}) \right| \left| \mathcal{L}(h(\mathbf{x}), f_1^t(\mathbf{x})) \right| d\mathbf{x} \\
    &\leq \epsilon_1^s(h) + \mu \mathbb{E}_{\mathcal{D}_1^s}\left[\left| f^t_1(\mathbf{x}_{1}^s) - f^s_1(\mathbf{x}_{1}^s) \right|\right] + \mu \int \left|\mathcal{D}_1^{t}(\mathbf{x}) - \mathcal{D}_1^s(\mathbf{x}) \right| \left| h(\mathbf{x}) - f_1^t(\mathbf{x}) \right| d\mathbf{x} \\
    &\leq \epsilon_1^s(h) + \mu \mathbb{E}_{\mathcal{D}_1^s}\left[\left| f^t_1(\mathbf{x}_{1}^s) - f^s_1(\mathbf{x}_{1}^s) \right|\right] + \mu d_1(\mathcal{D}^s_1, \mathcal{D}^t_1)
\end{align*}
Then we can obtain
\begin{align*}
    \epsilon_1^t(h) &\leq \epsilon_1^s(h) + \mu d_1(\mathcal{D}^s_1, \mathcal{D}^t_1) + \mu \min\left\{\mathbb{E}_{\mathcal{D}_1^s}\left[\left| f^t_1(\mathbf{x}_{1}^s) - f^s_1(\mathbf{x}_{1}^s) \right|\right], \mathbb{E}_{\mathcal{D}_1^t}\left[\left| f^t_1(\mathbf{x}_{1}^s) - f^s_1(\mathbf{x}_{1}^s) \right|\right] \right\} \\
    &=\epsilon_1^s(h) + \mu d_1(\mathcal{D}^s_1, \mathcal{D}^t_1) + \mu \lambda_*(\mathcal{D}^s_1, \mathcal{D}^t_1)
\end{align*}
where $\mathcal{D}_1^s$ and $\mathcal{D}_1^t$ are the density functions of $\mathcal{D}^s_1$ and $\mathcal{D}^t_1$, respectively. It is easy to see that this holds for source or target tasks across time stamps. For simplicity, we let $\Tilde{d} =\mu \max\left( \max_{1\leq j\leq N-1}d_1(\mathcal{D}^s_j, \mathcal{D}^s_{j+1}), d_1(\mathcal{D}^s_1, \mathcal{D}^t_1), \max_{1\leq j\leq N}d_1(\mathcal{D}^t_j, \mathcal{D}^t_{j+1}) \right)$, $\Tilde{\lambda} =\mu \max\left( \max_{1\leq j\leq N-1}\lambda_*(\mathcal{D}^s_j, \mathcal{D}^s_{j+1}), \lambda_*(\mathcal{D}^s_1, \mathcal{D}^t_1), \max_{1\leq j\leq N}\lambda_*(\mathcal{D}^t_j, \mathcal{D}^t_{j+1}) \right)$ and $\Tilde{\Re}(\mathcal{H}_{\mathcal{L}})$ is the Rademacher complexity of the class of functions $\mathcal{H}_{\mathcal{L}} = \{(\mathbf{x}, y) \mapsto \mathcal{L}(h(\mathbf{x}), y): h\in \mathcal{H} \}$, i.e.,  $\Tilde{\Re}(\mathcal{H}_{\mathcal{L}}) = \frac{1}{2N}\sum_{j=1}^N \left({\Re}_{D_j^s}(\mathcal{H}_{\mathcal{L}}) + {\Re}_{D_j^t}(\mathcal{H}_{\mathcal{L}}) \right) $.

\begin{equation*}
    \begin{aligned}
        \mathbb{E}_{\mathcal{B}} \left[g(\mathcal{B})\right] &= \mathbb{E}_{\mathcal{B}} \left[ \sup_{h\in \mathcal{H}} \epsilon_{{N+1}}^t(h) - \frac{1}{2N} \sum_{j=1}^N \left(\hat{\epsilon}_j^s(h) + \hat{\epsilon}_j^t(h) \right) \right] \\
        &= \mathbb{E}_{\mathcal{B}} \left[ \sup_{h\in \mathcal{H}} \epsilon_{{N+1}}^t(h) - \frac{1}{2N} \sum_{j=1}^N \left({\epsilon}_j^s(h) + {\epsilon}_j^t(h) \right) + \frac{1}{2N} \sum_{j=1}^N \left({\epsilon}_j^s(h) - \hat{\epsilon}_j^s(h) \right) + \frac{1}{2N} \sum_{j=1}^N \left({\epsilon}_j^t(h) - \hat{\epsilon}_j^t(h) \right)  \right] \\
        &= \frac{1}{2N} \sup_{h\in \mathcal{H}} \left(\sum_{j=1}^N \left(\epsilon_{{N+1}}^t(h) - \epsilon_j^t(h)\right) + \sum_{j=1}^N \left(\epsilon_{{N+1}}^t(h) - \epsilon_j^s(h)\right) \right) \\
        &\quad + \mathbb{E}_{\mathcal{B}} \left[ \sup_{h\in \mathcal{H}} \frac{1}{2N} \sum_{j=1}^N \left({\epsilon}_j^s(h) - \hat{\epsilon}_j^s(h) \right) + \frac{1}{2N} \sum_{j=1}^N \left({\epsilon}_j^t(h) - \hat{\epsilon}_j^t(h) \right)  \right] \\
        &\leq \frac{1}{2N} \sup_{h\in \mathcal{H}} \left(\sum_{j=1}^N \left(\epsilon_{{N+1}}^t(h) - \epsilon_j^t(h)\right) + \sum_{j=1}^N \left(\epsilon_1^s(h) - \epsilon_j^s(h)\right) + \sum_{j=1}^N \left(\epsilon_{{N+1}}^t(h) - \epsilon_1^t(h)\right) + \sum_{j=1}^N \left(\epsilon_1^t(h) - \epsilon_1^s(h)\right)\right) \\
        &\quad + \mathbb{E}_{\mathcal{B}} \left[ \frac{1}{2N} \sum_{j=1}^N \sup_{h\in \mathcal{H}} \left({\epsilon}_j^s(h) - \hat{\epsilon}_j^s(h) \right) + \frac{1}{2N} \sum_{j=1}^N \sup_{h\in \mathcal{H}} \left({\epsilon}_j^t(h) - \hat{\epsilon}_j^t(h) \right)  \right] \\
        &\leq \frac{1}{2N} \left( \frac{N(N+1)}{2}\left( \Tilde{d} + \Tilde{\lambda} \right) + \frac{N(N-1)}{2}\left( \Tilde{d} + \Tilde{\lambda} \right) + N \left( \Tilde{d} + \Tilde{\lambda} \right) + N \left( \Tilde{d} + \Tilde{\lambda} \right) \right) \\
        &\quad+ \mathbb{E}_{\mathcal{B}} \left[ \frac{1}{2N} \sum_{j=1}^N \hat{\Re}_{D_j^s}(\mathcal{H}_{\mathcal{L}}) + \frac{1}{2N} \sum_{j=1}^N \hat{\Re}_{D_j^t}(\mathcal{H}_{\mathcal{L}}) \right] \\
        &\leq \frac{N+2}{2} \left( \Tilde{d} + \Tilde{\lambda} \right) + \Tilde{\Re}(\mathcal{H}_{\mathcal{L}})
    \end{aligned}
\end{equation*}

Therefore, for any $h\in \mathcal{H}$,
\begin{align*}
        \epsilon_{{N+1}}(h) & \leq \frac{1}{2N} \sum_{j=1}^N \left(\hat{\epsilon}_j^s(h) + \hat{\epsilon}_j^t(h) \right) + \mathbb{E}_{\mathcal{B}} \left[g(\mathcal{B})\right] + \frac{\mu}{N}\sqrt{\frac{\log{\frac{1}{\delta}}}{2\Tilde{m}}} \\
        &\leq \frac{1}{2N} \sum_{j=1}^N \left(\hat{\epsilon}_j^s(h) + \hat{\epsilon}_j^t(h) \right) + \frac{N+2}{2} \left( \Tilde{d} + \Tilde{\lambda} \right) + \Tilde{\Re}(\mathcal{H}_{\mathcal{L}}) + \frac{\mu}{N}\sqrt{\frac{\log{\frac{1}{\delta}}}{2\Tilde{m}}}
\end{align*}
which completes the proof.
\end{proof}

\subsection{Proof of Corollary~\ref{col:marginal}}\label{sec:cor_mul}
Corollary~\ref{col:marginal} assume that the loss function $\mathcal{L}(\cdot, \cdot)$ is $\mu$-admissible and symmetric (i.e., $\mathcal{L}(y_1, y_2) = \mathcal{L}(y_2, y_1) $ for $y_1,y_2\in \mathcal{Y}$), and obeys the triangle inequality. Then
\begin{enumerate}[label=(\alph*), leftmargin=*,noitemsep,nolistsep]
\item when using {\bf $f$-divergence}~\cite{pmlr-v139-acuna21a}, denoted by $d_f$, the error bound of Theorem~\ref{continuous_transfer_learning_bound} holds with
\begin{align*}
    \Tilde{d} &= \max\big\{\max_{1\leq j\leq N-1}d_f(\mathcal{D}^s_j, \mathcal{D}^s_{j+1}), d_f(\mathcal{D}^s_1, \mathcal{D}^t_1), \max_{1\leq j\leq N}d_f(\mathcal{D}^t_j, \mathcal{D}^t_{j+1}) \big\} \\
    \Tilde{\lambda} &= \max\big\{\max_{1\leq j\leq N-1} \lambda_*(\mathcal{D}^s_j, \mathcal{D}^s_{j+1}), \lambda_*(\mathcal{D}^s_1, \mathcal{D}^t_1), \max_{1\leq j\leq N}\lambda_*(\mathcal{D}^t_j, \mathcal{D}^t_{j+1}) \big\}
\end{align*}
where $\lambda_*(\mathcal{D}^s_1, \mathcal{D}^t_1) = \min_{h\in \mathcal{H}} \epsilon_1^s(h) + \epsilon_1^t(h)$. Note that we have similar error bounds when using other marginal domain discrepancy measures (e.g., $\mathcal{H}$-divergence~\cite{ben2010theory}, discrepancy distance~\cite{mansour2009domain}, or MMD~\cite{gretton2012kernel}).

\item when using {\bf $\mathcal{C}$-divergence}~\cite{wu2020continuous} (measuring the distribution discrepancy over joint data distribution on $\mathcal{X}\times \mathcal{Y}$), denoted by $d_{\mathcal{C}}$, the error bound of Theorem~\ref{continuous_transfer_learning_bound} holds with
\begin{align*}
    \Tilde{d} &=\mu\cdot \max\big\{\max_{1\leq j\leq N-1}d_{\mathcal{C}}(\mathcal{D}^s_j, \mathcal{D}^s_{j+1}), d_{\mathcal{C}}(\mathcal{D}^s_1, \mathcal{D}^t_1), \max_{1\leq j\leq N}d_{\mathcal{C}}(\mathcal{D}^t_j, \mathcal{D}^t_{j+1}) \big\} \\
    \Tilde{\lambda} &= 0
\end{align*}
\end{enumerate}

\begin{proof}
It can be proven using the following observations. 

(a) When $f$-divergence is used, it can be proven by combining with the generalization bound of Theorem 2 in~\cite{pmlr-v139-acuna21a}.

In addition, we have similar error bounds when using other marginal domain discrepancy measures (e.g., $\mathcal{H}$-divergence~\cite{ben2010theory}, discrepancy distance~\cite{mansour2009domain}, or MMD~\cite{gretton2012kernel}).
\begin{enumerate}[label=(\roman*)]
\item using {\bf $\mathcal{H}$-divergence}, the error bound of Theorem~\ref{continuous_transfer_learning_bound} holds with 
\begin{align*}
    \Tilde{d} &=\mu\cdot \max\big\{\max_{1\leq j\leq N-1}d_{\mathcal{H}\Delta\mathcal{H}}(\mathcal{D}^s_j, \mathcal{D}^s_{j+1}), d_{\mathcal{H}\Delta\mathcal{H}}(\mathcal{D}^s_1, \mathcal{D}^t_1),\max_{1\leq j\leq N}d_{\mathcal{H}\Delta\mathcal{H}}(\mathcal{D}^t_j, \mathcal{D}^t_{j+1}) \big\} \\
    \Tilde{\lambda} &= \max\big\{\max_{1\leq j\leq N-1} \lambda_*(\mathcal{D}^s_j, \mathcal{D}^s_{j+1}), \lambda_*(\mathcal{D}^s_1, \mathcal{D}^t_1), \max_{1\leq j\leq N}\lambda_*(\mathcal{D}^t_j, \mathcal{D}^t_{j+1}) \big\}
\end{align*}
where $\lambda_*(\mathcal{D}^s_1, \mathcal{D}^t_1) = \min_{h\in \mathcal{H}} \epsilon_1^s(h) + \epsilon_1^t(h)$.

\item using {\bf discrepancy distance}, the error bound of Theorem~\ref{continuous_transfer_learning_bound} holds with
\begin{align*}
    \Tilde{d} &= \max\big\{\max_{1\leq j\leq N-1}d_{\mathrm{disc}}(\mathcal{D}^s_j, \mathcal{D}^s_{j+1}), d_{\mathrm{disc}}(\mathcal{D}^s_1, \mathcal{D}^t_1), \max_{1\leq j\leq N}d_{\mathrm{disc}}(\mathcal{D}^t_j, \mathcal{D}^t_{j+1}) \big\} \\
    \Tilde{\lambda} &= \max\big\{\max_{1\leq j\leq N-1} \lambda_*(\mathcal{D}^s_j, \mathcal{D}^s_{j+1}), \lambda_*(\mathcal{D}^s_1, \mathcal{D}^t_1),  \max_{1\leq j\leq N}\lambda_*(\mathcal{D}^t_j, \mathcal{D}^t_{j+1}) \big\}
\end{align*}
where $\lambda_*(\mathcal{D}^s_1, \mathcal{D}^t_1) = \mathbb{E}_{\mathcal{D}_1^s}\left[\mathcal{L}\left(h^*_{s_1}(\mathbf{x}), f_1^s(\mathbf{x})\right)\right] + \mathbb{E}_{\mathcal{D}_1^t}\left[ \mathcal{L} \left(h^*_{t_1}(\mathbf{x}), f_1^t(\mathbf{x})\right) \right] + \mathbb{E}_{\mathcal{D}_1^t}\left[\mathcal{L}\left(h^*_{t_1}(\mathbf{x}), h^*_{s_1}(\mathbf{x}) \right) \right]$, $h^*_{s_j}$ ($h^*_{t_j}$) is the optimal hypothesis for the $j^{\text{th}}$ source (target) task, and $L>0$ is a constant.

\item using {\bf Maximum Mean Discrepancy (MMD)}, the error bound of Theorem~\ref{continuous_transfer_learning_bound} holds with
\begin{align*}
    \Tilde{d} &= L\cdot \max\big\{\max_{1\leq j\leq N-1}d_{\mathrm{disc}}(\mathcal{D}^s_j, \mathcal{D}^s_{j+1}), d_{\mathrm{disc}}(\mathcal{D}^s_1, \mathcal{D}^t_1), \max_{1\leq j\leq N}d_{\mathrm{disc}}(\mathcal{D}^t_j, \mathcal{D}^t_{j+1}) \big\} \\
    \Tilde{\lambda} &= \max\big\{\max_{1\leq j\leq N-1} \lambda_*(\mathcal{D}^s_j, \mathcal{D}^s_{j+1}), \lambda_*(\mathcal{D}^s_1, \mathcal{D}^t_1),  \max_{1\leq j\leq N}\lambda_*(\mathcal{D}^t_j, \mathcal{D}^t_{j+1}) \big\}
\end{align*}
where $\lambda_*(\mathcal{D}^s_1, \mathcal{D}^t_1) = \mathbb{E}_{\mathcal{D}_1^s}\left[\mathcal{L}\left(h^*_{s_1}(\mathbf{x}), f_1^s(\mathbf{x})\right)\right] + \mathbb{E}_{\mathcal{D}_1^t}\left[ \mathcal{L} \left(h^*_{t_1}(\mathbf{x}), f_1^t(\mathbf{x})\right) \right] + \mathbb{E}_{\mathcal{D}_1^t}\left[\mathcal{L}\left(h^*_{t_1}(\mathbf{x}), h^*_{s_1}(\mathbf{x}) \right) \right]$, $h^*_{s_j}$ ($h^*_{t_j}$) is the optimal hypothesis for the $j^{\text{th}}$ source (target) task, and $L>0$ is a constant.

When $\mathcal{H}$-divergence is used, let $h^* = \arg\min_{h\in \mathcal{H}} \epsilon_1^s(h) + \epsilon_1^t(h)$ and $\lambda_*(\mathcal{D}^s_1, \mathcal{D}^t_1) = \min_{h\in \mathcal{H}} \epsilon_1^s(h) + \epsilon_1^t(h)$, it holds that
\begin{align*}
    \epsilon_1^t(h) &\leq \epsilon_1^t(h^*) + \epsilon_1^t(h,h^*) \\
    &\leq \epsilon_1^t(h^*) + \epsilon_1^s(h,h^*) + \left|\epsilon_1^t(h,h^*) - \epsilon_1^s(h,h^*) \right|\\
    &\leq \epsilon_1^t(h^*) + \epsilon_1^s(h^*) + \epsilon_1^s(h) + \mu \left|\int \left(\mathcal{D}_1^{t}(\mathbf{x}) - \mathcal{D}_1^s(\mathbf{x}) \right)  \left| h(\mathbf{x}) - h^*(\mathbf{x}) \right| d\mathbf{x} \right|\\
    &\leq \epsilon_1^s(h) + \mu\cdot d_{\mathcal{H}\Delta\mathcal{H}}(\mathcal{D}^s_1, \mathcal{D}^t_1) + \lambda_*(\mathcal{D}^s_1, \mathcal{D}^t_1)
\end{align*}
where $\mathcal{D}_1^s$ and $\mathcal{D}_1^t$ are the density functions of $\mathcal{D}^s_1$ and $\mathcal{D}^t_1$, respectively.

When discrepancy distance or MMD is used, based on Theorem 8 in \cite{mansour2009domain}, it holds that
\begin{align*}
    \epsilon_1^t(h)
    &\leq \epsilon_1^s(h) + d_{\mathrm{disc}}\left(\mathcal{D}^s_1, \mathcal{D}^t_1\right) + \mathbb{E}_{\mathcal{D}_1^s}\left[\mathcal{L}\left(h^*_{s_1}(\mathbf{x}), f_1^s(\mathbf{x})\right)\right] + \mathbb{E}_{\mathcal{D}_1^t}\left[ \mathcal{L} \left(h^*_{t_1}(\mathbf{x}), f_1^t(\mathbf{x})\right) \right] + \mathbb{E}_{\mathcal{D}_1^t}\left[\mathcal{L}\left(h^*_{t_1}(\mathbf{x}), h^*_{s_1}(\mathbf{x}) \right) \right]\\
    &= \epsilon_1^s(h) + d_{\mathrm{disc}}(\mathcal{D}^s_1, \mathcal{D}^t_1) + \lambda_*(\mathcal{D}^s_1, \mathcal{D}^t_1)
\end{align*}
where $\lambda_*(\mathcal{D}^s_1, \mathcal{D}^t_1) = \mathbb{E}_{\mathcal{D}_1^s}\left[\mathcal{L}\left(h^*_{s_1}(\mathbf{x}), f_1^s(\mathbf{x})\right)\right] + \mathbb{E}_{\mathcal{D}_1^t}\left[ \mathcal{L} \left(h^*_{t_1}(\mathbf{x}), f_1^t(\mathbf{x})\right) \right] + \mathbb{E}_{\mathcal{D}_1^t}\left[\mathcal{L}\left(h^*_{t_1}(\mathbf{x}), h^*_{s_1}(\mathbf{x}) \right) \right]$. Based on the lemma of Domain scatter bounds discrepancy in~\cite{ghifary2016scatter}, when the loss function $\mathcal{L}(\cdot, \cdot)$ is the squared loss, there exists a constant $L > 0$, it holds:
\begin{equation*}
d_{\mathrm{disc}}(\mathcal{D}, \mathcal{D}') \leq L\cdot d_{\mathrm{MMD}}(\mathcal{D}, \mathcal{D}')
\end{equation*}
where $d_{\mathrm{MMD}}(\cdot,\cdot)$ is defined on an reproducing kernel Hilbert space (RKHS) with a universal kernel.
\end{enumerate}

(b) Following Theorem 3.4 in \cite{wu2020continuous}, it holds that
\begin{align*}
    \epsilon_1^t(h) &= \epsilon_1^s(h) + \epsilon_1^t(h) -  \epsilon_1^s(h) \\
    &\leq \epsilon_1^s(h) + \mu \left|\int \left(\mathcal{D}_1^{t}(\mathbf{x}) - \mathcal{D}_1^s(\mathbf{x}) \right)  \left| h(\mathbf{x}) - y \right| d\mathbf{x}dy \right| \\
    &= \epsilon_1^s(h) + \mu \cdot d_{\mathcal{C}}(\mathcal{D}^s_1, \mathcal{D}^t_1)
\end{align*}
which completes the proof.
\end{proof}

\subsection{Pseudocode of \model}\label{sec:Pseudocode}
The overall training procedures of \model{} are summarized in Algorithm~\ref{alg:algorithm}. It is given dynamic source tasks with adequate labeled examples and dynamic target tasks with only unlabeled examples as input, and outputs the predicted class-labels for examples from the new target task $\mathcal{D}_{N+1}^t$. We first reformulate the dynamic tasks into a set of meta-pairs of consecutive tasks (Step 1), and then sequentially optimize the prior model initialization as well as generate the pseudo-labels for unlabeled examples from the target task (Steps 2-6). The learned model parameters can then be efficiently adapted to the newest task with only a few updates (Step 7).
\begin{algorithm}[tb]
\caption{Learning to Evolve (\model)}
\label{alg:algorithm}
\textbf{Input}: A labeled dynamic source task $\{\mathcal{D}_j^s\}_{j=1}^N$ and an unlabeled dynamic target task $\{\mathcal{D}_j^t\}_{j=1}^N$, the newest target task $\mathcal{D}_{N+1}^t$.\\
\textbf{Output}: Class labels on the new task $\mathcal{D}_{N+1}^t$.\\\vspace{-3mm}
\begin{algorithmic}[1] 
\STATE Create meta-pairs of tasks with $\{\mathcal{D}_j^s\}_{j=1}^N$ and $\{\mathcal{D}_j^t\}_{j=1}^N$;

$---------$ {\bf Meta-training} $---------$

\FOR{$k=0$ to $N$}
\STATE Learn the model initialization $\Tilde{\theta}^*$ via Eq. (\ref{eq:meta-train});
\STATE Fine-tune on unlabeled data of task $\mathcal{D}_{k+1}^t$ via Eq.(\ref{eq:inner_update});
\STATE Generate the pseudo-label for $\mathcal{D}_{k+1}^t$;
\ENDFOR

$---------$ {\bf Meta-testing} $---------$

\STATE Fine-tune on the newest target task $\mathcal{D}_{N+1}^t$ via Eq.(\ref{eq:meta-test});
\STATE \textbf{return} Predicted labels on the newest target task $\mathcal{D}_{N+1}^t$.
\end{algorithmic}
\end{algorithm}

\subsection{More Discussion}\label{sec:discussion}
In this work, we assume that the source and target tasks are continuously evolving over time. In other words, the data distribution of the source or target task is similar at the adjacent time stamps. This naturally motivates us to design the meta-pairs of tasks using adjacent tasks. However, it is possible that there exist other related tasks at the non-adjacent time stamps. One extreme case is that when the evolution of task distribution is negligible, it will be close to the transfer learning scenarios with multiple sources~\cite{zhao2018adversarial,wen2020domain}. In this case, any two historical tasks can be considered as the meta-pair of tasks in our \model\ framework. 

Note that compared with continuous transfer learning algorithms~\cite{bobu2018adapting,wang2020continuously,liu2020learning}, our proposed \model\ framework has the following benefits. (1) It leverages the evolution knowledge from both historical source and target tasks. (2) It could be efficiently adapted to the newest target task with just a few updates. (3) It mitigates the catastrophic forgetting on historical target tasks by learning the prior model initialization shared across meta-pairs of tasks (that is, it could preserve the predictive performance on historical tasks by fine-tuning on those tasks with a few updates). (4) It is flexible to incorporate any existing static transfer learning algorithms~\cite{pan2009survey} by instantiating $\zeta(\cdot)$ of Eq. (\ref{eq:meta-train}) accordingly.

\subsection{Additional Results}\label{sec:additional_results}
To better reproduce our experiments, we provide the following details of experimental setting as well as some additional experimental results. All the experiments are performed on a Windows machine with four 3.80GHz Intel Cores, 64GB RAM and one NVIDIA Quadro RTX 5000 GPU. The source code can be found in our supplemental materials.

\begin{table}[t]
    \centering
    \begin{tabular}{lcccc}
    \toprule
         & \# examples/task & \# classes & \# source time stamps & \# target time stamps \\ \midrule
      Office-31  & up to 2800 & 31 & 5 & 6 \\
      Image-CLEF & $\sim$600 & 12 & 5 & 6 \\
      Caltran & $\sim$480 & 2 & 1 & 11 \\
    \bottomrule
    \end{tabular}
    \vspace{-3mm}
    \caption{Summary of data sets used in the experiments}
    \label{tab:dataset}
\end{table}

\begin{table}[t]
    \centering
    \begin{tabular}{lccccccc}
    \toprule
         & Batch size & Outer epochs & Outer learning rate & Inner epochs & Inner learning rate & $\gamma$ \\\midrule
        Office-31 & 32 & 20 & 0.5 & 1 & 0.5 & 0.1 \\
        Image-ClEF & 48 & 5 & 0.01 & 5 & 1.0 & 0.1 \\
        Caltran & 32 & 10 & 0.001 & 30 & 0.1 & 0.1 \\
    \bottomrule
    \end{tabular}
    \vspace{-3mm}
    \caption{Reproducible hyperparameter settings}
    \label{tab:hyperparams}
\end{table}

{\bf Data summary:} We used three publicly available image data sets: Office-31\footnote{\url{https://people.eecs.berkeley.edu/~jhoffman/domainadapt/}} (with 3 tasks: Amazon, Webcam and DSLR), Image-CLEF\footnote{\url{http://imageclef.org/2014/adaptation/}} (with 4 tasks: B, C, I and P) and Caltran\footnote{\url{http://cma.berkeleyvision.org}}. For Office-31 and Image-CLEF, there are 5 time stamps in the source task and 6 time stamp in the target task. Caltran contains the real-time images captured by a camera at an intersection for several weeks. Table~\ref{tab:dataset} summarizes the statistics of the image data sets used in the experiments. In this case, we assume all the tasks in one data set will share the labeling space, thus enabling the knowledge transfer from source task to a set of time evolving target tasks when no labeled training examples are available within any target task. However, the first two data sets consist of only static tasks. Thus, we generate a set of time evolving task by adding the random noise and rotation to the original images. Take Office-31 (Amazon $\rightarrow$ Webcam) as an example, on one hand, we generate the dynamic labeled source task $\mathcal{D}_j^s (j=1,2,\cdots,5)$ at time stamp $j$ from Amazon by rotating the original images with degree $O_d$ (i.e., $O_d =-30\cdot (j-1)$) and adding the random Gaussian noise. On the other hand, we generate the dynamic unlabeled target task $\mathcal{D}_j^t (j=1,2,\cdots,6)$ at time stamp $j$ from Webcam by rotating the original images with degree $O_d'$ and adding the random salt\&pepper noise with magnitude $O_n'$, i.e., $O_d' =15\cdot (j-1)$ and $O_n' =0.05\cdot (j-1)$. The goal of dynamic transfer learning is to learn the predicted class labels for the newest target task $\mathcal{D}_6^t$ using historical source and target knowledge. Caltran is a real-world image data set captured by a camera at an intersection for several weeks, so we consider that one task consists of the images from a single day. In this case, following~\cite{hoffman2014continuous}, we take the first day's images as the labeled source task, and others as a dynamic unlabeled target task.

{\bf Hyper-parameters:} We adopted the ResNet-18~\cite{he2016deep} pretrained on ImageNet as the base network for feature extraction with an added 256-dimension bottleneck layer between the {\em res5c} and {\em fc} layers. For fair comparison, we use the same feature extraction for all baseline models. In addition, we set $\gamma=0.1$ and $p=80$ for all the experiments.
Table~\ref{tab:hyperparams} shows the hyper-parameters used in our experiments for different data sets. For inner optimization that updates the model parameters on each meta-pair of tasks given the model initialization, we adopt stochastic gradient descent (SGD) with mini-batch where the batch size is 32 or 48. In addition, we set the number of validation images as 32 in every task, and then use all other images as the training set.

{\bf Convergence and Efficiency:} 
We discuss the convergence and computational complexity of \model\ as follows. Following~\cite{fallah2020convergence}, it holds that model-agnostic meta-learning (MAML)~\cite{finn2017model} finds an $\epsilon'$-first-order stationary point for any $\epsilon'>0$ after at most $O(1/\epsilon'^2)$ iterations under mild conditions. Moreover, if the inner learning rate $\alpha$ is small, the approximation error of its first-order approximation (FO-MAML) induced by ignoring the second-order term would not impact its convergence. Therefore, our \model\ framework inherits this convergence theory, as we optimize the model initialization of Eq. (\ref{eq:meta-train}) via gradient-based MAML methods. In addition, the computational complexity of \model\ will be linear in the number of trainable parameters $\theta$ when FO-MAML is adopted in our experiments. More specifically, we also empirically compare the running cost of \model\ with other baselines. It is observed that the computational costs of TransLATE and DARN on B$\rightarrow$P are 114.18 (seconds) and 295.14 (seconds), respectively. As a comparison, our \model\ needs 234.63 (seconds) for meta-training and only 11.81 (seconds) for fine-tuning on meta-testing. This is consistent with our analysis in Subsection~\ref{sec:discussion} that \model\ could be efficiently adapted to the newest target task with just a few update during meta-testing.

\end{document}